\documentclass[sn-mathphys-num,Numbered,iicol]{sn-jnl}

\usepackage{graphicx}%
\usepackage{amsfonts}%
\usepackage{amsthm}%
\usepackage{mathrsfs}%
\usepackage[title]{appendix}%
\usepackage{xcolor}%
\usepackage{textcomp}%
\usepackage{manyfoot}%
\usepackage{algorithm}%
\usepackage{algorithmicx}%
\usepackage{algpseudocode}%
\usepackage{listings}%
\usepackage{amsmath}
\usepackage{amssymb}
\usepackage{multicol}
\usepackage{multirow}
\usepackage{makecell}
\usepackage{color}
\usepackage{pifont}
\usepackage{abbreviation}
\usepackage{booktabs}
\usepackage{anyfontsize}
\usepackage{bm}

\definecolor{sqColor}{rgb}{0.0,0.0,0.0}

\definecolor{green}{rgb}{0.0,1.0,0.0}
\newcommand{\green}[1]{{\color{green} #1}}
\definecolor{red}{rgb}{1.0,0.0,0.0}
\newcommand{\red}[1]{{\color{red} #1}}

\newcommand{\cmark}{\green{\ding{51}}}%
\newcommand{\xmark}{\red{\ding{55}}}

\begin{document}

\title[FineSkiing: A Fine-grained Benchmark for Skiing Action Quality Assessment]{\centering FineSkiing: A Fine-grained Benchmark for Skiing Action Quality Assessment}


\author[1]{\fnm{Yongji} \sur{Zhang}}\email{zhangyongji1998@gmail.com}
\equalcont{These authors contributed equally to this work.}
\author[2]{\fnm{Siqi} \sur{Li}}\email{lisiqi19971013@gmail.com}
\equalcont{These authors contributed equally to this work.}
\author[2]{\fnm{Yue} \sur{Gao}}\email{kevin.gaoy@gmail.com}
\author[1*]{\fnm{Yu} \sur{Jiang}}\email{jiangyu2011@jlu.edu.cn}

\affil[\small 1]{\small  \orgdiv{\{Key Laboratory of Symbolic Computation and Knowledge Engineering of Ministry of Education, College of Computer Science and Technology\}}, \orgname{Jilin University}, \orgaddress{\city{Changchun}, \postcode{130012}, \country{China}}}
\affil[\small 2]{\small  \orgdiv{\{BNRist, THUIBCS, BLBCI, School of Software\}}, \orgname{Tsinghua University}, \orgaddress{\city{Beijing}, \postcode{100084}, \country{China}}}

\abstract{
Action Quality Assessment (AQA) aims to evaluate and score sports actions, which has attracted widespread interest in recent years. Existing AQA methods primarily predict scores based on features extracted from the entire video, resulting in limited interpretability and reliability. Meanwhile, existing AQA datasets also lack fine-grained annotations for action scores, especially for deduction items and sub-score annotations. In this paper, we construct the first AQA dataset containing fine-grained sub-score and deduction annotations for aerial skiing, which will be released as a new benchmark. For the technical challenges, we propose a novel AQA method, named JudgeMind, which significantly enhances performance and reliability by simulating the judgment and scoring mindset of professional referees. Our method segments the input action video into different stages and scores each stage to enhance accuracy. Then, we propose a stage-aware feature enhancement and fusion module to boost the perception of stage-specific key regions and enhance the robustness to visual changes caused by frequent camera viewpoints switching. In addition, we propose a knowledge-based grade-aware decoder to incorporate possible deduction items as prior knowledge to predict more accurate and reliable scores. Experimental results demonstrate that our method achieves state-of-the-art performance.
}

\keywords{Action Quality Assessment, Fine-grained Benchmark, Deduction Annotations, Knowledge-based Modeling}

\maketitle

\section{Introduction}
\label{sec:intro}
As the key technology for analyzing and evaluating competitive sports actions, Action Quality Assessment (AQA) has attracted significant attention in recent years, \eg, the Judging Support System (JSS) was leveraged in the Paris Olympics for gymnastics to efficiently and accurately help judges to evaluate actions and reduce disputes in subjective scoring, resulting in fair and accurate scoring.

Existing AQA methods~\cite{ActionNet,cofinal,CoRe,gdlt,finediving,MS-LSTM,SVR,TPT,fineparser,logo,pan2021adaptive,li2022pairwise,xu2024finesports} mainly segment the input video based on fixed length, \eg, splitting the input video into clips of 32 frames and extracting features. However, the actual sports action often contains several distinct stages, \eg, diving contains approach, takeoff, dive, and rip entry, while aerial skiing contains air, form, and landing. Professional judges need to evaluate the completion of actions within each stage and the continuity between actions for scoring. Therefore, segmenting the video into a fixed length cannot achieve effective action assessment. To address this issue, FineDiving~\cite{finediving} leverages a segmentation module to segment each sub-action, \eg, splitting each turn or flip in the ``dive'' stage of diving. However, in practice, the intervals between sub-actions may not be clear due to the limitations of frame rate and capturing viewpoint. The semantic similarity of contiguous sub-actions leads to insufficient sub-action segmentation accuracy. In addition, the relationship between sub-actions is crucial for action assessment, \eg, ``timing'' is an important deduction item that reflects the overall smoothness of the action. Therefore, segmenting the video by sub-action is also not in compliance with the scoring standard.

\begin{figure}
\begin{center}
\includegraphics[width=1.0\linewidth]{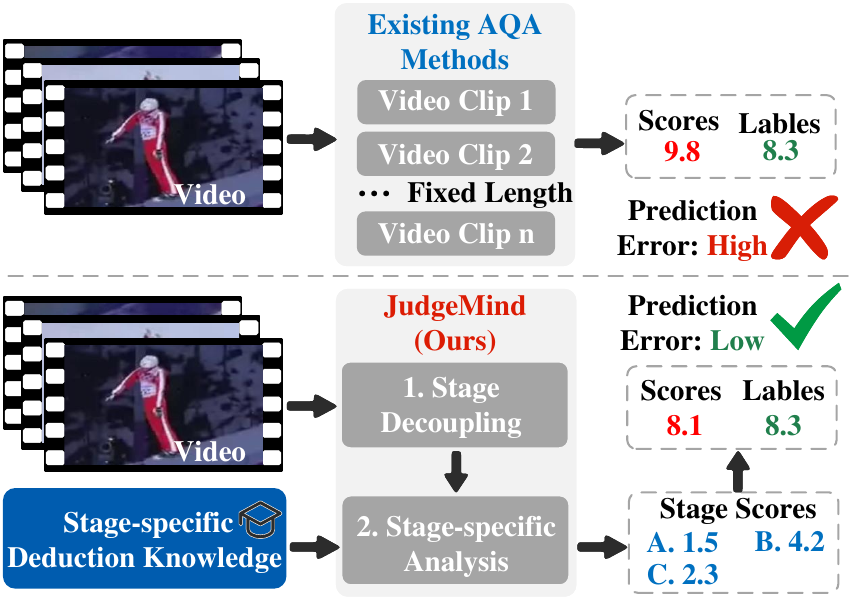}
\end{center}
\caption{
\textbf{Comparison of existing AQA methods with our proposed method.} Our proposed method simulates the scoring mindset of professional judges for more accurate and reliable scoring.}
\label{fig:1}
\end{figure}

On the other hand, existing AQA methods directly extract visual features based on different backbone networks, and regress the action scores based on the features. However, this is unclear for action quality assessment since the final score cannot be interpreted by analyzing the quality of each action stage, leading to low reliability and interpretability. For competitive sports, \eg, gymnastics and skiing, professional judges will deduct scores for each stage of the athlete's action according to standardized deduction rules. Meanwhile, the key points of deduction are different in each stage, \eg, for skiing, the ``form'' stage focuses on the action posture of the athlete, while the ``air'' and ``landing'' stages focus on the spatial relationship between the athlete and the course. Existing AQA methods lack of fully utilizing these prior knowledge, resulting in insufficient performance.

In addition, existing AQA datasets, \eg, FineDiving~\cite{finediving} and RG~\cite{ActionNet}, contain only the final scores of the entire action and lack fine-grained annotations including stage scores and deduction reasons, \etc, which is also a principal impediment to the advancement of research within this domain. 

To address the technical challenges, we propose a novel AQA method that simulates the professional \textbf{judge}'s scoring \textbf{mind}set, named \textbf{JudgeMind}, as shown in Fig.~\ref{fig:1}. Taking video frames as input, our method first uses a temporal segmentation module to segment the input video into different stages for separate processing, which is more compatible with the scoring process of human judges. Then, we use a stage-aware feature extraction module to extract different features for different stages, allowing the model to accurately focus on the key regions of each stage. This module also tackles the changes due to the switching of near and far camera views. Subsequently, the features are enhanced based on the athlete's action code prior and are forwarded into our proposed Knowledge-based Grade-aware Decoder (K-GAD). K-GAD encodes the possible deduction items of each stage as knowledge and then interacts with the video features, allowing our proposed method to learn stage-specific deduction knowledge and evaluate the action quality more accurately. Finally, scores for each stage are generated based on Likert Scoring and are summed to obtain the final complete action score. Our proposed method simulates the mindset of professional judges and incorporates knowledge of action codes and deductions to achieve precise and reliable AQA results. 

\begin{figure}
\begin{center}
\includegraphics[width=1.0\linewidth]{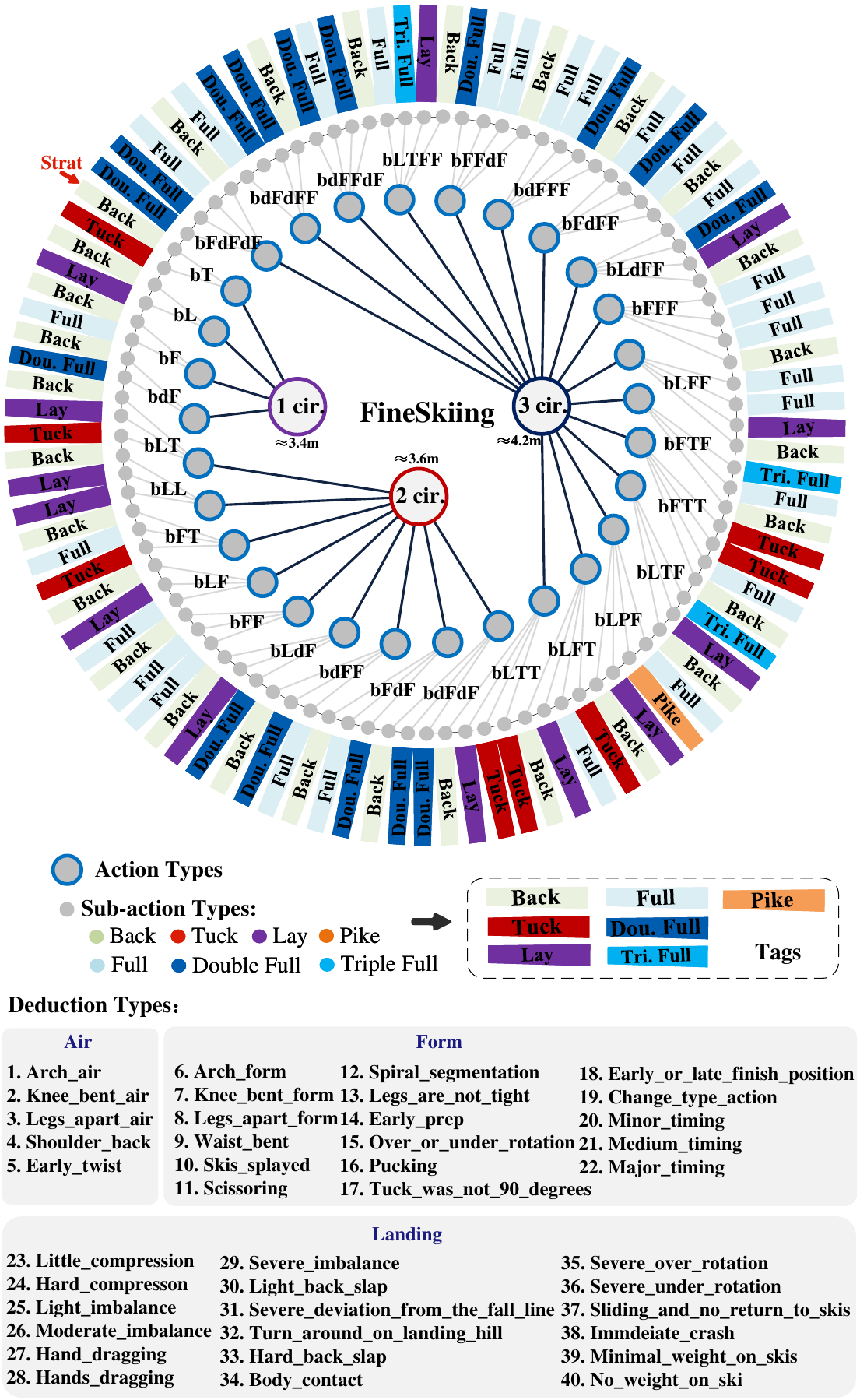}
\caption{
List of action/sub-action types and stage-specific deductions items contained in our FineSkiing dataset. 
}
\end{center}
\label{fig:2}
\end{figure}

To address the data challenges, we construct the first AQA dataset containing stage scores and detailed deduction annotations based on aerial skiing, named \textbf{FineSkiing}, as shown in Fig.~\ref{fig:2}. Specifically, we collected videos of the men's and women's aerial skiing from the last three Winter Olympics, as well as training videos from national teams to expand the data diversity. The dataset is precisely labeled in strict compliance with the International Ski Federation scoring rules and verified by Olympic-qualified judges to ensure that the annotations are accurate. Compared with existing AQA datasets, our dataset contains fine-grained stage scores and deduction annotations for the first time.

Our contribution could be summarized as follow:
\begin{itemize}
    \item We propose a novel AQA framework, named JudgeMind, that simulates the professional judge's scoring mindset by stage-decoupling and integrating deduction knowledge to assess scores for each action stage.
    \item We propose a stage-aware action feature extraction module that achieves effective action representation based on foreground and global feature extraction and fusion. We propose a knowledge-based grade-aware decoder that encodes deduction knowledge and interacts with action features to achieve accurate action assessment.
    \item We construct a large-scale fine-grained AQA dataset, named \textbf{FineSkiing}, which contains detailed sub-scores and deduction annotations of each stage for the first time\footnote{Our code and dataset will be released after acceptance.}.
\end{itemize}

\section{Related Work}
\label{sec: related work}
\subsection{Action Quality Assessment}
\label{sec:sai}
Action Quality Assessment (AQA) methods are now widely applied in sports action evaluation~\cite{shao2020finegym,ramanathan2014human,kong2022human,sportshhi,sun2022human} and scoring~\cite{wang2021survey,zahan2024learning,limagr,gao2023automatic,zeng2024multimodal}. Early AQA methods~\cite{SVR,baller,pan2019action,venkataraman2015dynamical} are mainly based on handcrafted features to estimate video scores. More recently, various deep learning AQA models~\cite{ActionNet,gdlt,CoRe,cofinal,finediving,TPT,rica,finedving+} are proposed, using convolutional networks, graph networks, recurrent networks, and Transformers as backbone. Parmar~\etal~\cite{MIT} leverage spatiotemporal features to capture the dynamics of actions like diving, vault gymnastics, and figure skating. Xu~\etal~\cite{MS-LSTM} use self-attention mechanisms and multi-scale dilated convolutional LSTMs to aggregate information, achieving improved performance. Pan~\etal~\cite{HGCN} utilize graph neural networks to model joint relationships, providing a more accurate assessment of visual action performance. Meanwhile, some exemplar-based AQA methods are proposed to evaluate actions by comparing input videos with reference videos of the same action type and score. Tang~\etal~\cite{UNLV} introduce USDL, which learns uncertainty-aware score distributions to reduce the inherent ambiguity in human judges' scoring labels. Yu~\etal~\cite{CoRe} develop a contrastive regression framework (CoRe) based on video-level features, ranking videos and predicting accurate scores. GDLT~\cite{gdlt} adopted a Likert-scale approach for grading action quality, while CoFInAL~\cite{cofinal} further enhanced this method's performance and interpretability using a hierarchical approach. 

Although existing methods achieve favorable AQA performance, they do not fully utilize the judging rules and standards of action evaluation, resulting in poor robustness and reliability. In contrast, our proposed method simulates the mindset of human judges and integrates action code and deduction knowledge prior, enabling more accurate and interpretable action quality assessment.

\subsection{Spots Video Datasets}

Several sports video datasets are released for the AQA task. Pirsiavash~\etal~\cite{MIT} construct the first Olympic judging dataset, which included the MIT-Dive dataset and the MIT-Skate dataset, focusing on diving and figure skating. Based on this work, Xu~\etal~\cite{MS-LSTM} construct the UNLV Dive and UNLV Vault datasets, which consist of 370 diving videos and 176 gymnastics videos, respectively. Then, the AQA-7 dataset~\cite{action} is constructed, containing 7 types of sports and a total of 1,189 videos. MTL-AQA~\cite{MTL} is a larger AQA dataset with 1,412 videos across 16 different sports. Fis-V~\cite{MS-LSTM} is specifically developed for figure skating containing 500 videos. RG~\cite{ActionNet} focused on rhythmic gymnastics, containing 1,000 videos. More recently, Xu~\etal~\cite{finediving} construct the first fine-grained AQA dataset FineDiving, providing detailed sub-action annotations. 

However, the annotation of existing AQA datasets is still inadequate and lacks more detailed annotations of sub-scores and deduction items, which is necessary for accurate and reliable AQA.

\section{The FineSkiing Dataset}
\subsection{Dataset Construction}

\begin{figure*}[!htbp]
\begin{center}
\includegraphics[width=1.0\linewidth]{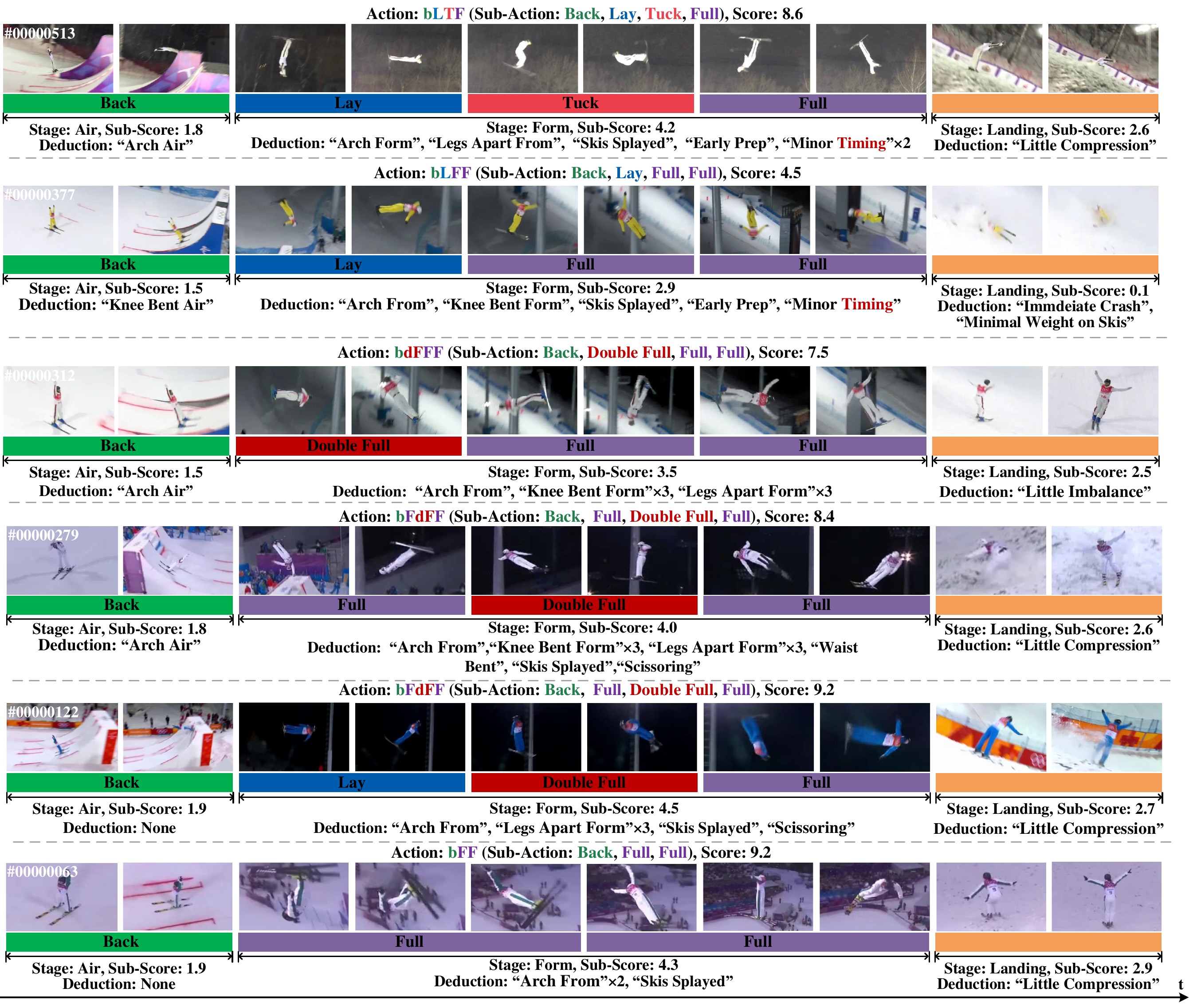}
\end{center}

\caption{\textbf{Examples from our FineSkiing dataset.} Each row shows the video frames of an entire aerial skiing maneuver. The action type and overall score are annotated above the snapshot. Sub-scores of each stage (air, form, and landing) and the detailed deductions are shown below. It should be noted that deductions have detailed temporal location labeling.}
\label{fig:3}
\end{figure*}

To tackle the above-mentioned lack of datasets, we construct a \textbf{Fine}-grained AQA dataset based on aerial \textbf{Skiing} that containing sub-score of each stage and detailed deduction items, named \textbf{FineSkiing}. Our FineSkiing dataset contains video clips collected from the qualification and final rounds of the men's and women's aerial skiing events at the 2014, 2018, and 2022 Winter Olympics. Moreover, since the athletes in the Olympics mainly perform similarly complex maneuvers, \eg, Back Full-Full-Double Full (bFFdF), we additionally collect training videos of national teams qualified for the Olympics, thereby forming our comprehensive and diversified dataset. Most of our data is gathered from the International Ski Federation (FIS), which includes official videos of top international competitions and training videos.

\begin{table}[htbp!]
\footnotesize
\centering
\caption{\textbf{Comparison of our FineSkiing dataset with existing AQA datasets.} \textbf{Score} represents the total score of each maneuver. \textbf{Act.} represents temporal boundary annotation of each sub-action. \textbf{Stage} represents the temporal boundary annotation of each action stage. \textbf{Stage Score} represents the sub-score of each stage. \textbf{Dedu. Item} represents the deduction item annotations.}  
\label{tab:01}

\setlength{\tabcolsep}{0.8mm}
\begin{tabular}{l|cccccc}
\toprule
\textbf{Dataset}                     & \textbf{Score} & \textbf{Act.}   & \textbf{Stage}    & \textbf{\makecell[c]{Stage\\Score}}   & \textbf{\makecell[c]{Dedu.\\Item}}    \\ \midrule 
\textbf{MIT Olympic~\cite{MIT}}      & \cmark & \xmark & \xmark & \xmark & \xmark \\
\textbf{UNLV Olympic~\cite{UNLV}}    & \cmark & \xmark & \xmark & \xmark & \xmark \\
\textbf{AQA-7~\cite{action}}         & \cmark & \xmark & \xmark & \xmark & \xmark \\
\textbf{MTL-AQA~\cite{MTL}}          & \cmark & \xmark & \xmark & \xmark & \xmark \\ 
\textbf{Fis-V~\cite{MS-LSTM}}        & \cmark & \xmark & \xmark & \xmark & \xmark \\ 
\textbf{RG~\cite{ActionNet}}         & \cmark & \xmark & \xmark & \xmark & \xmark \\
\textbf{FS1000~\cite{FS1000}}        & \cmark & \xmark & \xmark & \xmark & \xmark \\ \midrule 
\textbf{FineDiving~\cite{finediving}}& \cmark & \cmark & \xmark & \xmark & \xmark \\ \midrule
\textbf{FineSkiing}                  & \cmark & \cmark & \cmark & \cmark & \cmark \\ \bottomrule
\end{tabular}

\end{table}

For dataset processing, we extracted the full sequences of aerial skiing, ensuring that each clip contains the entire maneuver from starting to landing. The resolution of the videos is standardized to 1080p or 720p, and the frame rate is unified at 30 FPS to ensure high-quality temporal analysis. For dataset annotation, we refer to existing AQA datasets, including FineDiving~\cite{finediving} and RG~\cite{ActionNet}, and annotate each video clip with detailed metadata, including:
\begin{itemize}
\item Gender of the athlete.
\item Action type, \ie, the jump code of specific maneuver.
\item Degree of difficulty (predefined based on the maneuver).
\item Final score, which is calculated based on the scores of 5 official judges, with the highest and lowest scores discarded.
\item Sub-action boundary, \ie, the starting and ending timestamp of each sub-action.
\item \textbf{Stage boundary.} For the first time, the starting and ending timestamp of the three key stages in aerial skiing, \ie, air, form, and landing, are provided.
\item \textbf{Stage score}. For the first time, detailed sub-score of each stage are provided.
\item \textbf{Deduction}. For the first time, detailed deduction reasons and the timestamp of each deduction are provided.
\end{itemize}

The annotation process \textbf{strictly follows the FIS scoring guidelines} and is \textbf{validated by FIS-certified judges qualified for the Olympics judgment}. This ensures that the final scores and deduction annotations are both accurate and aligned with professional judging standards.

\subsection{Dataset Statistics}

Our FineSkiing dataset contains 550 videos, among which 440 videos are selected as the training set, and the remaining 110 videos are selected as the test set. Each video has an average duration of 7.7 seconds and is detailed annotated with the temporal boundary of each action stage and sub-action, total score, the sub-score of each stage, and fine-grained deduction items. 

Figure~\ref{fig:3} shows several examples of our FineSkiing dataset, where each athlete's complete performance is segmented into three key stages, \ie, the ``air'' (takeoff) stage, the ``form'' (complex action) stage, and the ``landing'' (touchdown) stage. Each maneuver is composed of specific sub-actions, such as the `` bLTF (back-lay-tuck-full)'' sequence shown in the first row. As shown in each row of Fig.~\ref{fig:3}, we annotate the overall score and the sub-scores of each stage. Additionally, similar to the existing FineDiving dataset~\cite{finediving}, we also label the start and end times of each sub-action. For the first time, we annotate all deductions for the entire maneuver in detail, including the deduction items and the corresponding temporal location, as shown in Fig.~\ref{fig:3}. 

In addition to deduction items, we also annotated the severity of deductions based on skiing scoring standards, including:
\begin{itemize}
    \item Minor Deductions,
    \item Medium Deductions,
    \item Major Deductions,
    \item Absolute Deductions.
\end{itemize}

Even for the same penalty, the deductions can vary depending on the maneuver's difficulty coefficient and the severity of the error. As shown in the first and second rows of Fig.~\ref{fig:3}, two athletes incurred similar penalties during their routines. However, they received significantly different deductions due to differing maneuver difficulties and error magnitudes. Similarly, in the third and fourth rows of Fig.~\ref{fig:3}, both athletes committed an ``Arch Air" error during the ``air" stage. However, since the first athlete's error is more severe, he is penalized with a larger deduction score. For more details, please refer to the supplementary materials and the publicly available dataset we will release shortly.

Table~\ref{tab:01} shows the comparison of our FineSkiing dataset with existing AQA datasets, including MIT Olympic~\cite{MIT}, UNLV Olympic~\cite{UNLV}, AQA-7~\cite{action}, MTL-AQA~\cite{MTL}, RG~\cite{ActionNet}, Fis-V~\cite{MS-LSTM}, FS1000~\cite{FS1000}, and the existing fine-grained AQA dataset FineDiving~\cite{finediving}. From the table, we could observe that our FineSkiing dataset achieves a significant improvement in fine-grained annotations compared to previous datasets. The comprehensive annotations and detailed deduction descriptions enable more accurate, interpretable, and robust AQA model training. Specifically, compared to other AQA datasets, our FineSkiing dataset offers several notable advantages:
\begin{itemize}
\item It is the first dataset focused on aerial skiing, offering rich data tailored for this highly technical sport.
\item It is the first AQA dataset that contains detailed deduction annotations, providing both qualitative and quantitative descriptions of deductions, which is crucial for transparent and explainable predictions.
\item It is strictly annotated based on the FIS official judging handbook, ensuring that all data matches professional assessment standards. This makes FineSkiing not only suitable for the AQA task but also for tasks such as temporal action localization and temporal action detection.
\end{itemize}

\section{Method}
\label{sec:method}
\begin{figure*}[t]
    \centering
    \includegraphics[width=1.0\linewidth]{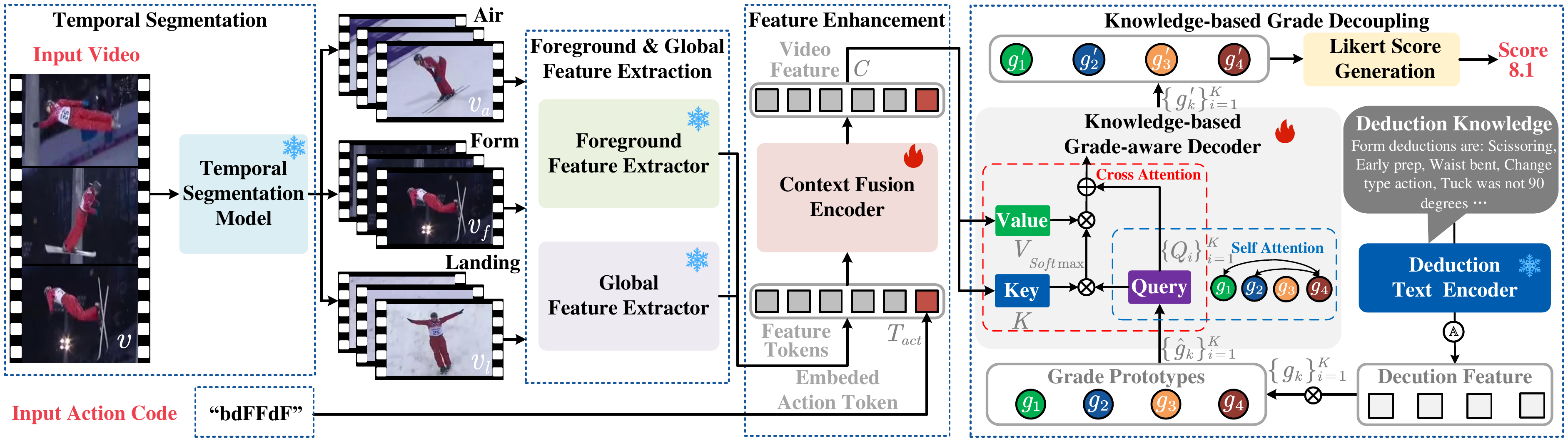}

     \caption{\textbf{Overview of our proposed method JudgeMind.} Taking action video as input, it is first segmented into three different stages by a temporal segmentation model. Then, a stage-aware feature extraction module is leveraged to extract stage-specific key region features, which are further fed into the context fusion encoder along with the athlete action code prior to obtain enhanced features. Subsequently, a knowledge-based grade-aware decoder is proposed to assess action grades by interacting motion features with deduction knowledge. Finally, the action score is calculated using a Likert scoring module. (Best viewed in color.)
    } %

    \label{fig:architecture}
\end{figure*}

In this section, we introduce our proposed method, named JudgeMind, a novel AQA method that is inspired by the mindset of professional judges. Figure~\ref{fig:architecture} shows the overall framework of our proposed method. We first introduce the overview of our proposed method in Sec.~\ref{sec:4.1}. Then, in Sec.~\ref{sec:4.2}, we introduce the detailed network architecture. Finally, in Sec.~\ref{sec:4.3}, we introduce the loss function used in our method.

\subsection{Overview}
\label{sec:4.1}

For vanilla AQA methods~\cite{ActionNet,gdlt,CoRe,cofinal,finediving,TPT,rica}, the input video $v$ is divided into fixed-length (\eg, 32 frames) clips $\{v_t\}_{t=1}^T$ using handcraft hyperparameters, and features are extracted separately from each clip. However, this paradigm ignores the semantic connections between clips and the semantic differences across sub-actions. In each stage, the possible deductions and the major focuses are different, \eg, deductions in the ``air" stage include ``knee bent air'' and ``legs apart air", \etc, and the corresponding points should be deducted accordingly. This leads to the vanilla AQA paradigm being unreliable.

Figure~\ref{fig:architecture} shows the overall pipeline of our proposed method. To overcome the above-mentioned challenge, our proposed method firstly segments the input video $v$ into different stages, \eg, air $v_{a}$, form $v_{f}$, and landing $v_{l}$ for skiing, using a \textbf{temporal segmentation model}. Therefore, each stage could be processed according to its unique characteristics and judging criteria. Then, the segmented videos are forwarded into our proposed \textbf{stage-aware feature extraction} module to extract features, which could extract stage-specific key region features and handle the dramatic visual changes caused by frequent camera view switching. Subsequently, the extracted features and the encoded action description prior are fed into the \textbf{context fusion encoder} to obtain enhanced features $C$ for each stage. The enhanced feature $C$ is then forwarded into the \textbf{knowledge-based grade-aware decoder} together with the stage-specific deduction knowledge to obtain grade-aware features, where the stage-specific deduction information is introduced to enhance the interpretability and refine the scoring precision. Finally, the grade-aware features of each stage are used to generate the final stage score using a Likert scoring module.

\subsection{Network Architecture}
\label{sec:4.2}

\textbf{Temporal Segmentation.} As mentioned above, our method leverages a temporal segmentation model to segment input video into distinct stages. In practice, we leverage the CLIP vision transformer~\cite{clip} as the temporal segmentation model. Specifically, for the input video $v$ containing $T$ frames, we use the segmentation model to extract features and use a linear layer to predict the stage of each input frame. It should be noted that some existing methods~\cite{finediving,finedving+} segment input video according to sub-actions. However, for sports action assessment, the association between sub-actions is critical since ``timing" is a key deduction item of action assessment. This paradigm will naturally corrupt the association between sub-actions.

\textbf{Stage-aware Feature Extraction.} Since the action videos may have frequent camera switching, and the key regions are different at different stages, directly extracting global features from each frame may not be effective for the AQA task. Therefore, we divide video frames into two types, \ie, core and non-core. For the non-core frames, \eg, the frames in the ``air" and ``landing" stages for skiing, the video typically captures the athlete’s overall movement trajectory, with the athlete occupying a smaller portion of the frame. Meanwhile, the deductions mainly depend on the spatial interactions between athletes and the course. For the core frames, the video mainly records the precise actions of the athlete, who tends to occupy a larger portion of the frame. Meanwhile, the deductions mainly depend on the athlete’s action form and quality. Therefore, background in these frames is basically interfering information.

To address these challenges, we propose a stage-aware feature extraction module that uses a global extractor and a foreground extractor to obtain the global information of the frame and the foreground information of the athlete, respectively, targeting the global spatial interactions and the athlete's fine actions. In practice, we discriminate the core and non-core videos based on stages. For non-core frames, both foreground and global features are extracted, while only foreground features are extracted for core frames. In practice, we use a pre-trained MixFormer~\cite{mixformer,mixformer2} tracker to detect the foreground athlete, and use a Swin Transformer~\cite{videoswin} encoder pre-trained on the Kinetics dataset~\cite{kinetics} to extract foreground features. For global features, we use an I3D~\cite{I3D} encoder pre-trained on ImageNet~\cite{imagenet} dataset to extract features from the entire frame. 

\textbf{Context Fusion Encoder.} After the global and foreground features are extracted, we leverage a context fusion encoder to achieve feature enhancement based on fine-grained action information. Specifically, for non-core frames, we use a ViT encoder~\cite{vit} to integrate the foreground and global features, representing the spatial interactions between athlete and course. For core frames, we first encode the action description text $T_{act}$, \eg, the code ``bdFFdF" represents the action ``back-double full-full-double full", by a CLIP text encoder~\cite{clip}. This allows the model to learn the prior knowledge of the athlete's maneuver, which is similar to the fact that the human referee will evaluate the athlete's actions based on the knowledge of what maneuver will be accomplished. The encoded action prior knowledge is then combined with the foreground features and fed into the ViT encoder~\cite{vit} to obtain the enhanced features $C$ focused on the athlete’s detailed actions.

\textbf{Knowledge-based Grade-aware Decoder.} After the enhanced features are obtained, we propose a knowledge-based grade-aware decoder (K-GAD) to predict the grade level of the action based on deduction knowledge. Specifically, we first introduce a set of $K$ learnable grade prototypes $\{g_{k}\}_{i=1}^{K}$. Each prototype represents a distinct action grade level, which allows the model to efficiently learn the unique characteristics of each grade level. Then, we encode the deduction prior knowledge of each stage, \eg, deduction items ``Pucking" and ``Early Prep" in the ``form'' stage, using BERT~\cite{bert} to obtain the deduction feature $F_{de}$, which is then multiplied with the grade prototypes to obtain deduction knowledge-based prototypes $\{\hat g_{k}\}_{i=1}^{K}$.

Then, the action grade level is predicted based on the interaction between the grade prototypes and video features. Specifically, the interaction is achieved using a Transformer decoder~\cite{vit}. The self-attention is calculated across the prototypes to explore the correlation between different grade levels, allowing each prototype to better represent each grade level. Meanwhile, grade-decoupled cross-attention is calculated across prototypes and video features, allowing each prototype to extract relevant action grade information. In practice, we leverage the video features ${C}$ and the knowledge-enhanced prototypes $\{\hat g_{k}\}_{i=1}^{K}$ as input. The query is first generated from the grade prototypes, and the key and value are generated from the video features using  three different linear projections:
\begin{equation}
    \mathbf{Q}_i= \mathbf{W}_q\hat{g}_i,  \mathbf{K}= \mathbf{W}_k C,  \mathbf{V}= \mathbf{W}_v C,
\end{equation}
where $\{\mathbf{Q}_{i}\}_{i=1}^{K}$,$\mathbf{K}$ and $\mathbf{V}$ indicate queries, key and value, respectively. Using the proposed K-GAD, the knowledge-based grade-aware features $\{g^\prime_{k}\}_{i=1}^{K}$ could be obtained.

\textbf{Likert Scoring.} Finally, we leverage the Likert scoring module~\cite{gdlt,likert2,cofinal} to predict the action score based on the grade-aware features $\{g^\prime_{k}\}_{i=1}^{K}$. Specifically, the action assessment score is generated through a linear combination of the grade-aware features and the grade values. It should be noted that our method predicts scores of each stage according to stage-specific action prior and deduction knowledge, and the final score is the summarization of all stage scores.

\subsection{Loss Function}
\label{sec:4.3}

We train our model using the Mean Squared Error (MSE). Specifically, we calculate the MSE between the predicted score $s^\text{pre}$ and labeled score $s^\text{gt}$ for each stage. The total loss is the sum of the loss for each stage score, ensuring the model learns accurate predictions at each stage:
 \begin{equation}
\label{loss}
\mathcal{L} = \sum_{i=1}^T w_i \mathcal{L}_2(s_i^\text{pre},s_i^\text{gt}),
\end{equation}
where $T$ is the number of stages, $w_i$ is the trade-off hyper-parameter.

Using our proposed method, accurate action quality assessment could be achieved by simulating the judging mindset of professional judges. By incorporating action type prior and stage-specific possible deduction knowledge, the scores for each stage could be predicted, and the final complete AQA result could be further obtained.

\section{Experiments}
\label{sec: experiments}
To demonstrate the effectiveness of our proposed method, we conduct experiments on our FineSkiing dataset, which is the only AQA dataset containing fine-grained stage scores and deduction annotations. In Sec.~\ref{sec:5.1}, we introduce the detailed experimental settings. The comparison of our method and existing state-of-the-art methods are analyzed in Sec.~\ref{sec:5.2}. In Sec.~\ref{sec:5.3}, we conduct ablation experiments to prove the validity of each proposed module.

\subsection{Experimental Settings}
\label{sec:5.1}

\textbf{Evaluation Metrics}
We leverage two widely used evaluation metrics for the AQA task, \ie, Spearman’s Rank Correlation Coefficient~\cite{finediving,spearman1,spearman2,gdlt} (\textbf{SRCC}) and Relative $\ell_{2}$ Distance~\cite{cofinal,gdlt,l2} ($\bm{R{\ell_{2}}}$). SRCC measures the rank correlation between the predicted scores and ground truth scores. A higher SRCC indicates that the model is better at predicting the relative order of scores. ${R{\ell_{2}}}$ quantifies error by measuring the $\ell_{2}$ distance between the predicted score and ground truth score, normalized by the ground truth score.

\textbf{Implementation Details.} For our proposed method, we use the fixed official pretrained weights of the action description and deduction text encoder, and fine-tune the temporal segmentation module, and the global and foreground feature extractor separately. Then, these modules are fixed and the whole model is trained end-to-end. Our method is implemented in PyTorch and is trained on NVIDIA V100 GPUs. The optimizer is SGD~\cite{sgd} with a momentum of 0.9. The initial learning rate is set to 0.01 and is decayed gradually to 0.0001 using a cosine annealing schedule~\cite{sgdr}. The whole model is trained for 150 epochs with a batch size of 32. We apply a dropout rate of 0.3 to regularize the model and set the weight decay to 0.01. For Eq.~\eqref{loss}, we set $w_i=1$.

\subsection{Comparisons with State-of-the-Art Methods}
\label{sec:5.2}

\begin{table}[tbp!]
\footnotesize
\centering
\caption{Quantitative comparison results on our FineSkiing dataset. Best results are in \textbf{bold}, second-best results are \underline{underlined}. \textbf{Back.} represents the backbone.}
\label{tab:02}
\setlength{\tabcolsep}{0.6mm}{
\begin{tabular}{l|c|l|c|c}
\toprule
\textbf{Method}             & \textbf{Back.} & \textbf{Pub/Year} & \textbf{SRCC}($\uparrow$) & $\bm{R{\ell_{2}}(\downarrow)}$    \\ \midrule
\multicolumn{5}{l}{\textbf{Exemplar-based Method}} \\ \midrule
CoRe~\cite{CoRe}                & I3D               & ICCV 2021         & 0.830                     & 2.666                             \\
TPT~\cite{TPT}                 & I3D               & ECCV 2022         & 0.822                     & 2.502                             \\
TSA~\cite{finediving}                 & I3D               & CVPR 2022         & 0.878                     & 1.558    \\       
STSA~\cite{finedving+}                 & I3D               & IJCV 2024         &  0.880                    & 1.696    \\ 
 \midrule
\multicolumn{5}{l}{\textbf{Exemplar-free Method}}                                                                                   \\ \midrule
C3D+SVR~\cite{SVR}             & C3D               & CVPR 2017         & 0.801                     & 2.359                             \\
MS-LSTM~\cite{MS-LSTM}             & C3D               & TCSVT 2019        & 0.802                     & 2.685                             \\ \midrule
MS-LSTM~\cite{MS-LSTM}             & I3D               & TCSVT 2019        & 0.841                     & 2.020                             \\
ACTION-NET~\cite{ActionNet}          & I3D               & ACMMM 2020        & 0.875                     & 1.504                             \\
GDLT~\cite{gdlt}                & I3D               & CVPR 2022         & 0.877         & 1.590                             \\
CoFInAl~\cite{cofinal}             & I3D               & IJCAI 2024        & \underline{0.883}                     & 1.509                             \\ \midrule
MS-LSTM~\cite{MS-LSTM}             & VST               & TCSVT 2019        & 0.825                     & 2.378                             \\
ACTION-NET~\cite{ActionNet}          & VST               & ACMMM 2020        & 0.874                     & \underline{1.468}                 \\
GDLT~\cite{gdlt}                & VST               & CVPR 2022         & 0.876                     & 1.953                             \\
CoFInAl~\cite{cofinal}      & VST               & IJCAI 2024        & 0.878                     & 1.749                             \\ 
RICA$^{2}$~\cite{rica}          &VST                &ECCV 2024           & 0.857  & 1.930          \\ 
\midrule 
\textbf{Ours}               & VST               & -                 & \textbf{0.927}            & \textbf{0.933}                    \\ \bottomrule 
\end{tabular}}
\end{table}

Table~\ref{tab:02} shows the quantitative comparisons on our FineSkiing dataset. Specifically, we compare our method with existing state-of-the-art exemplar-based methods and exemplar-free methods. All methods are trained on our FineSkiing dataset for fair comparison. Specifically, exemplar-based methods rely on the additional video with the same action code as the target video and the corresponding score as input to serve as reference. These methods heavily depend on the similarity between the template video and the target video. From the table, we could observe that our proposed method outperforms the exemplar-based method STSA~\cite{finedving+}, achieving an improvement of 0.047 in SRCC and a decline of 0.763 in $R{\ell_{2}}$. On the other hand, compared with the exemplar-free methods, our proposed method also achieves state-of-the-art performance. Specifically, compared to our baseline method GDLT~\cite{gdlt}, our proposed method could improve the SRCC from 0.876 to 0.927 and reduce the $R{\ell_{2}}$ from 1.953 to 0.933. Compared with the second-best method CoFInAl~\cite{cofinal}, our method could improve the SRCC by 5.0\% and reduce the $R{\ell_{2}}$ by 38.2\%. It should be noted that exemplar-free methods could outperform exemplar-based methods on our FineSkiing dataset. This is due to the fact that the background and capture viewpoint of each video in our FineSkiing dataset may differ notably, even for the same difficult maneuvers. Existing datasets, \eg, the widely used FineDiving dataset~\cite{finediving} that focuses on diving videos, contain videos mainly captured from the same viewpoint (\eg, side view), making the template video more similar to the target video. However, exemplar-based methods may not be able to adaptively tackle the background and viewpoint differences between the template video and the target video, which limits their performance. These quantitative results demonstrate the effectiveness of our proposed method.

\begin{table}[tbp!]
\footnotesize
\centering
\caption{Comparison results on the existing FineDiving dataset~\cite{finediving}. Best results are in \textbf{bold}, second-best results are \underline{underlined}. It should be noted that the performance of our proposed method is hindered by the lack of stage-specific annotations and detailed deduction items in the FineDiving dataset. \textbf{Back.} represents the backbone.}
\label{tab:03}
\setlength{\tabcolsep}{0.6mm}{
\begin{tabular}{l|c|l|cc}
\toprule
\textbf{Method}             & \textbf{Back.} & \textbf{Pub/Year} & \textbf{SRCC}($\uparrow$) & $\bm{R{\ell_{2}}(\downarrow)}$    \\ \midrule
C3D+SVR~\cite{SVR}             & C3D               & CVPR 2017                                  &0.732 &0.988\\
MS-LSTM~\cite{MS-LSTM}             & C3D               & TCSVT 2019                                 &0.773 &0.862\\ \midrule
MS-LSTM~\cite{MS-LSTM}             & I3D               & TCSVT 2019                                  &0.790 &0.755\\
ACTION-NET~\cite{ActionNet}          & I3D               & ACMMM 2020                                     &0.857&0.528\\
GDLT~\cite{gdlt}                & I3D               & CVPR 2022                                    &0.866&0.583\\
CoFInAl~\cite{cofinal}      & I3D               & IJCAI 2024                                  &0.867 &0.681   \\ 
\midrule
MS-LSTM~\cite{MS-LSTM}             & VST               & TCSVT 2019                               &0.875&0.549     \\
ACTION-NET~\cite{ActionNet}          & VST               & ACMMM 2020                        &0.877&0.463\\
GDLT~\cite{gdlt}                & VST               & CVPR 2022                                   &0.909&\underline{0.362}\\
CoFInAl~\cite{cofinal}      & VST               & IJCAI 2024                                     &\underline{0.912}&0.368\\ 
\midrule 
\textbf{Ours}               & VST               & -                 & \textbf{0.924}          &   \textbf{0.321}               \\ \bottomrule 
\end{tabular}}
\end{table}

Table~\ref{tab:03} further shows the quantitative comparisons on the existing FineDiving~\cite{finediving} dataset. Specifically, we compared our method with existing state-of-the-art exemplar-free methods. For a fair comparison, all methods are trained on the FineDiving dataset. Our proposed method achieves state-of-the-art performance compared to other exemplar-free methods. Specifically, compared to our baseline method GDLT~\cite{gdlt}, our method improves the SRCC from 0.909 to 0.924 and reduces the $R{\ell_{2}}$ from 0.362 to 0.321. Additionally, compared to the second-best method CoFInAl~\cite{cofinal}, our proposed method could also improve the SRCC by 0.012.

\begin{table*}[htbp]
\centering
\footnotesize
\caption{Ablation studies on our FineSkiing dataset. \cmark denotes the component is added. \xmark denotes the component is removed.
}
\label{tab:04}

\setlength{\tabcolsep}{2.1mm}{
\begin{tabular}{@{}cccccccccccccc@{}}  
\toprule
&    &\multirow{2}{*}{\textbf{\makecell[c]{Temporal\\Segmentation}}} &&\multicolumn{2}{c}{\textbf{Feature Extraction}}   &&\multicolumn{2}{c}{\textbf{Action Knowledge}}     &&\multicolumn{2}{c}{\textbf{Final Score}}                        \\ \cline{5-6} \cline{8-9} \cline{11-12} 
&\rule{0pt}{8pt}    &                                               &&\textbf{Foreground}   &\textbf{Global}             &&\textbf{Action Code}  & \textbf{Deduction Know.}  && \textbf{SRCC($\uparrow$)}    &\bm{$R\ell_{2}(\downarrow)$}     \\ \midrule
&(1) & \xmark                                                        &&\xmark                &\cmark                     &&\xmark                &\xmark                     && 0.877                        & 1.590                           \\ \midrule
&(2) & \xmark                                                        &&\cmark                &\cmark                     &&\cmark                &\cmark                     && 0.908                        & 1.295                           \\ \midrule
&(3) & \cmark                                                        &&\cmark                &\xmark                     &&\cmark                &\cmark                     && 0.923                        & 0.949                           \\
&(4) & \cmark                                                        &&\xmark                &\cmark                     &&\cmark                &\cmark                     && 0.913                        & 1.007                           \\ \midrule 
&(5) & \cmark                                                        &&\cmark                &\cmark                     &&\cmark                &\xmark                     && 0.909                        & 1.245                           \\ 
&(6) & \cmark                                                        &&\cmark                &\cmark                     &&\xmark                &\cmark                     && 0.922                        & 0.944                           \\ \midrule
&(7) & \cmark                                                        &&\cmark                &\cmark                     &&\cmark                &\cmark                     && 0.927                        & 0.932                           \\ \bottomrule
\end{tabular}
}  
\end{table*}

\begin{table}
    \footnotesize
    \centering
    \caption{Ablation study of different loss ratios on our FineSkiing dataset.}
    \label{tab:05}
    \setlength{\tabcolsep}{2.7mm}
    \begin{tabular}{l|c|c|cc}
        \toprule
        $\mathcal{L}_\text{air}$ & $\mathcal{L}_\text{form}$ & $\mathcal{L}_\text{landing}$ &\textbf{SRCC}($\uparrow$) & $\bm{R{\ell_{2}}(\downarrow)}$\\        \midrule
        1&4&5 &0.919 &1.099 \\
        1&6&3 &0.918  &1.073 \\
        2&4&4 &0.918  &1.081  \\
        2&5&3 &0.921  &1.070  \\
        3&4&3 &0.921  &1.052\\
        \midrule
        1&1&1 &0.927  &0.933 \\
        \bottomrule
    \end{tabular}
\end{table}

It should be noted that the FineDiving dataset lacks fine-grained annotations for individual stages, we are constrained to use overall scores for training rather than stage-specific supervision. This limitation hinders the ability of our proposed method to optimize the performance at each stage, resulting in less effective training than on our FineSkiing dataset, which provides explicit stage-wise annotations. These detailed annotations enable the model to capture stage-specific characteristics more effectively. While our model can still be applied to the FineDiving dataset, the absence of critical stage-level information restricts its potential performance. In addition, the FineDiving dataset lacks detailed deduction items. Therefore, we manually collect the deduction knowledge as input for our method. This may also have impacts on model performance.

\begin{figure}
\begin{center}
\includegraphics[width=1.0\linewidth]{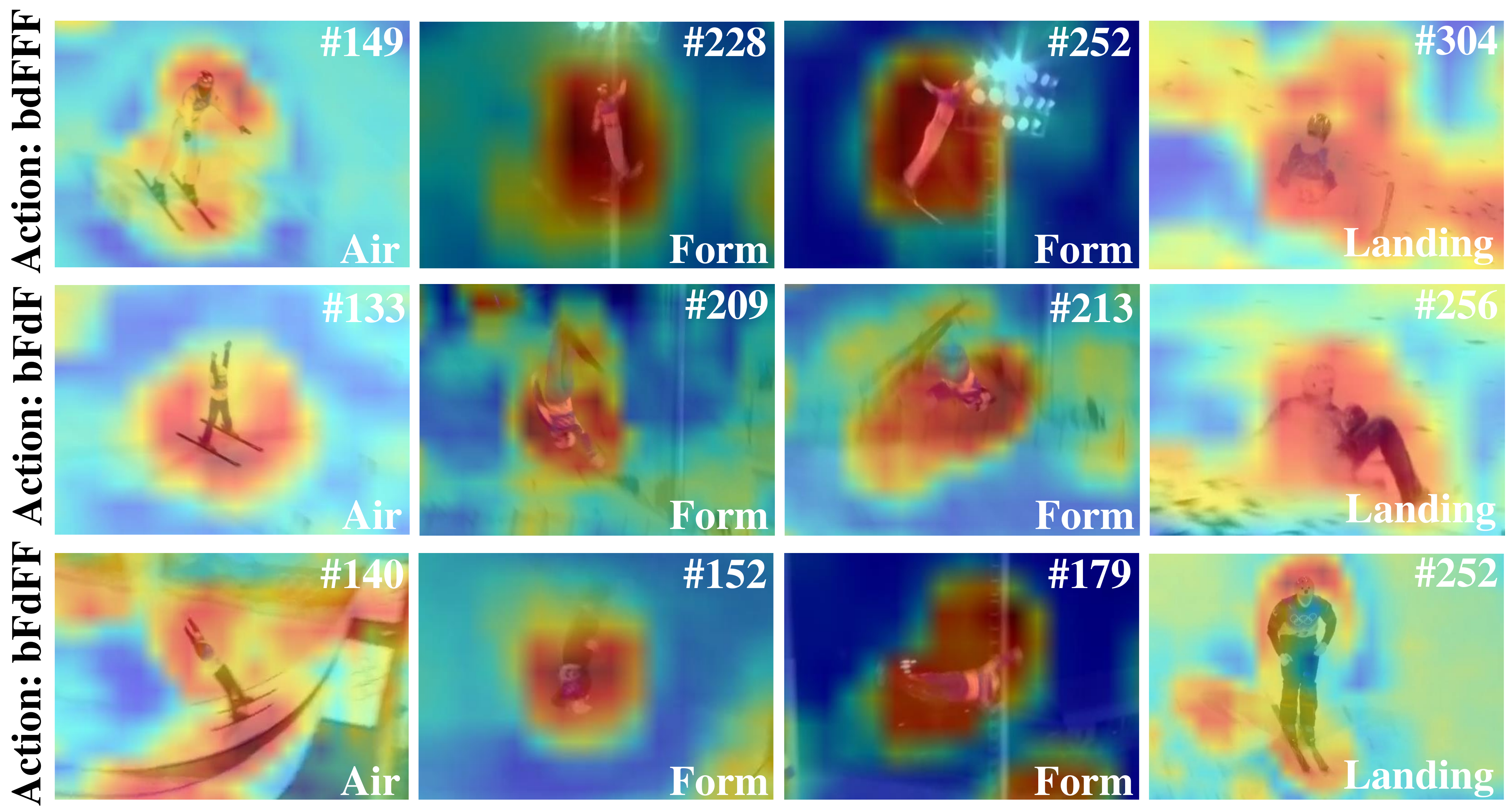}
\end{center}

\caption{\textbf{Feature heat maps at different stages.} For the ``air'' and ``landing'' stages, the model mainly focuses on the interaction between the athlete and the course, while in the ``form'' stage, it focuses on the maneuvers of the athlete.}
\label{fig:5}
\end{figure}

Figure~\ref{fig:5} and Fig.~\ref{fig:6} provide more qualitative results. Figure~\ref{fig:5} shows the feature heat maps at different stages. From the figure, we could observe that our model could focus on both the athlete and the course in the ``air'' and ``landing'' stages in order to assess the athlete posture and the interaction with the course. For the ``form'' stage, our model focuses accurately on the athlete to effectively assess the action posture. Figure~\ref{fig:6} further visualizes the correlation between the predicted scores $\hat{s}_i$ with the ground truth scores $s_i$. The dashed line denotes the linear regression equation $\hat{s}_i = k \cdot s_i + b$, where $k$ and $b$ are shown in the top-left corner. The solid line indicates the ideal case $\hat{s}_i = s_i$. The Spearman correlation coefficient is also shown above. The shaded region around the regression line represents the confidence interval. From the figure, we could observe that our method achieves the best performance with the highest Spearman coefficient. Furthermore, the regression line for our method is closest to the ideal orange line, indicating that the predicted scores are the most consistent with the ground truth scores.

\subsection{Ablation Study }
\label{sec:5.3}
To prove the validity of each proposed module, we conduct ablation experiments on our FineSkiing dataset, as shown in Tab.~\ref{tab:04}. Specifically, we validate the performance of our model with or without the temporal segmentation, foreground and global feature extraction, action code prior, and deduction knowledge, respectively. Our baseline model is GDLT~\cite{gdlt}, \ie, our full model with the removal of temporal segmentation module, foreground feature extraction, and action knowledge prior input, as shown in row (1) of Tab.~\ref{tab:04}.

\textbf{Temporal segmentation.} To verify the necessity of stage decoupling, we remove the temporal segmentation model and process the complete video directly. The performance under such setting is shown in row (2). Compared with the full model, \ie, row (7), the removal of stage decoupling will lead to a decline of 0.05 in SRCC. This is because the assessment of different stages has different key points, and the lack of stage decoupling will obscure semantics and lead to performance degradation.

\textbf{Foreground and global extraction.} Row (3) and row (4) of Tab.~\ref{tab:04} show the quantitative results of our proposed method with the removal of global and foreground feature extraction, respectively. 
From the table, we could observe that the removal of global and foreground feature extraction modules will lead to declines of 0.004 and 0.014 in SRCC, respectively. The experimental results also prove that foreground feature extraction plays a more critical role, which is intuitive since features of the athlete's region are more essential for action assessment. Meanwhile, global features are also indispensable.

\begin{figure}
\begin{center}
\includegraphics[width=1.0\linewidth]{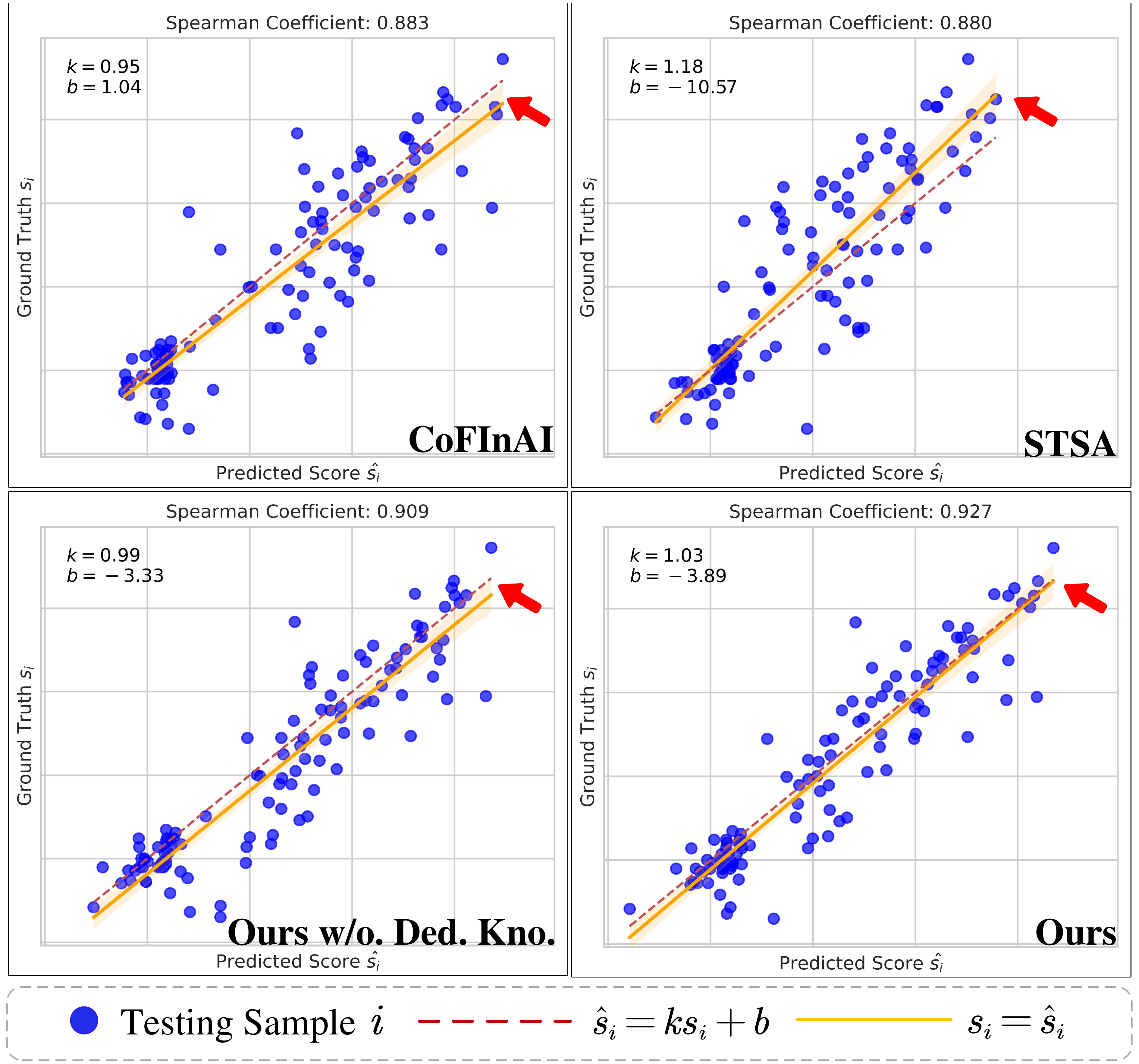}
\end{center}
\caption{Correlation between predicted and ground truth scores for different methods. (Best View in Color.)}
\label{fig:6}
\end{figure}

\textbf{Stage-specific action knowledge.} To prove the necessity of input action knowledge prior, we remove the input stage-specific deduction knowledge and action code, and the results are shown in row (5) and row (6), respectively. From the table, we could observe that the removal of input action code will decrease the SRCC by 0.005, and the removal of the deduction knowledge will lead to a decline of 0.018. Figure~\ref{fig:6} further proves the validity of the deduction knowledge. From the second row of the figure, we could observe that the predicted scores do not fit well with the ground truth scores (\ie, the dashed line is far from the solid line) after removing the input deduction item knowledge. These results prove the effectiveness of the input action and deduction knowledge. 

Table~\ref{tab:05} shows the model performance with different weight ratios of the loss functions in each stage. Different from other sports that focus primarily on the action quality of some specific stages, \eg, diving AQA focuses primarily on in the flight and entry stages, the quality of action in all stages of skiing is essential for AQA, with distinct key points and deductions respectively. Thus, from the table, we could observe that the best performance could be achieved with $w_i = 1$. 

\section{Conclusion}
In this paper, we propose a novel AQA method, named JudgeMind, that simulates the mindset of professional judges. Our proposed method segments input video according to action stages and processes each stage separately based on stage-specific knowledge. A stage-aware feature extraction module is leveraged to focus on distinct key regions of each stage, and a knowledge-based grade-aware decoder is proposed to effectively assess action grade levels based on stage-specific deduction knowledge. In addition, we construct the first fine-grained AQA dataset, named FineSkiing, containing detailed sub-scores and deduction annotations of each stage, which will be released as a new benchmark for this task. Experimental results prove that our method achieves state-of-the-art performance. We believe our work contributes technically to the reliable and robust AQA, and our dataset could support future research on stage-aware and fine-grained AQA task.

\section*{Data Availability Statements}
All experiments are based on the newly constructed FineSkiing dataset, which is ready to be released. We will release the FineSkiing dataset and our source code after the paper is accepted. Some example sequences could be downloaded from \url{https://drive.google.com/drive/folders/1RASpzn20WdV3uhZptDB-kufPG76W9FhH?usp=sharing} for review now.

\section*{Supplementary Material}
\addcontentsline{toc}{section}{Supplementary Material}


\section{Maneuver Code and Degree of Difficulty}


Table~\ref{tab:code} shows the correspondence between specific jump codes and the associated degree of difficulty (DD) in our FineSkiing dataset. The DD for male and female athletes performing the same maneuver may differ slightly. Sub-actions are connected using hyphens, and several sub-actions form a complete maneuver. In international competitions such as the Olympics, athletes often select triple-circle maneuvers. To enrich the dataset, we included training videos from Olympic-qualified teams. According to the International Ski Federation (FIS) scoring standards, five judges evaluate each stage of the maneuver. The final score is calculated by averaging the judges' scores (discard the highest and lowest scores) and multiplying the result by the maneuver's DD.

\begin{table}[htbp!]
\footnotesize
\centering
\caption{\textbf{Aerial Jump Code and Degree of Difficulty Chart.} The difficulty coefficients for males and females correspond to the movement codes shown in Figure 3 of the main text. All action based official off single kicker. 
``DD.'' denotes Degree of Difficulty, ``Dou.'' denotes Double.}  
\label{tab:code}
\setlength{\tabcolsep}{0.45mm}{
\begin{tabular}{l|c c c}
\toprule
\textbf{Jump Description}            & \textbf{Code} & \textbf{Men} & \textbf{Women} \\
\midrule
\multicolumn{4}{l}{\textbf{One Circle}}       \\ \midrule
Back Tuck                            & bT                 & 2.000            & 2.000             \\
Back Pike                            & bP                 & 2.000            & 2.000             \\
Back Lay                             & bL                 & 2.050            & 2.050             \\
Back Full                            & bF                 & 2.300            & 2.300             \\
Back Dou. Full                     & bdF                & 2.900            & 2.900             \\
\midrule
\multicolumn{4}{l}{\textbf{Two Circle}}       \\ \midrule
Back Tuck-Tuck                      & bTT               & 2.300            & 2.300             \\
Back Lay–Tuck                        & bLT                & 2.600            & 2.600             \\
Back Lay-Lay                         & bLL                & 2.650            & 2.650             \\
Back Full-Tuck                       & bFT                & 2.850            & 2.850             \\
Back Lay-Full                        & bLF                & 2.900            & 2.900             \\
Back Full-Full                       & bFF                & 3.150            & 3.150             \\
Back Dou. Full-Tuck                & bdFT               & 3.225            & 3.225             \\
Back Lay-Dou. Full                 & bLdF               & 3.275            & 3.275             \\
Back Dou. Full-Full                & bdFF               & 3.525            & 3.525             \\
Back Full-Dou. Full                & bFdF               & 3.525            & 3.525             \\
Back Lay-Triple Full                 & bLtF               & 3.750            & 3.750             \\
Back Dou. Full-Dou. Full         & bdFdF              & 3.900            & 3.900             \\
Back Full-Triple Full                & bFtF               & 4.000            & 4.000             \\
Back Triple Full-Full                & btFF               & 4.000            & 4.000             \\
\midrule
\multicolumn{4}{l}{\textbf{Three Circle}}       \\ \midrule
Back Lay-Tuck-Tuck                   & bLTT               & 3.200            & 3.392             \\
Back Lay-Full-Tuck                   & bLFT               & 3.500            & 3.710             \\
Back Lay-Pike-Full                   & bLPF               & 3.500            & 3.710             \\
Back Lay-Tuck-Full                   & bLTF               & 3.500            & 3.710             \\
Back Full-Tuck-Full                  & bFTF               & 3.750            & 3.975             \\
Back Lay-Full-Full                   & bLFF               & 3.800            & 4.028             \\
Back Full-Full-Full                  & bFFF               & 4.050            & 4.293             \\
Back Lay-Dou. Full-Full            & bLdFF              & 4.175            & 4.425             \\
Back Lay-Full-Dou. Full            & bLFdF              & 4.275            & 4.531             \\
Back Full-Dou. Full-Tuck           & bFdFT              & 4.125            & 4.373             \\
Back Full-Dou. Full-Full           & bFdFF              & 4.425            & 4.690             \\
Back Dou. Full-Full-Full           & bdFFF              & 4.525            & 4.796             \\
Back Full-Full-Dou. Full           & bFFdF              & 4.525            & 4.796             \\
Back Lay-Triple Full-Full            & bLtFF              & 4.650            & 4.929             \\
Back Dou. Full-Full-Dou. Full    & bdFFdF             & 5.000            & 5.300             \\
Back Dou. Full-Dou. Full-Full    & bdFdFF             & 5.100            & 5.406             \\
Back Full-Dou. Full-Dou. Full    & bFdFdF             & 5.100            & 5.406             \\
Back Full-Triple Full-Full           & bFtFF              & 5.200            & 5.512             \\
Back Full-Full-Triple Full           & bFFtF              & 5.300            & 5.618             \\
Back Dou. Full-Dou. Full-Dou. Full & bdFdFdF        & 5.675            & 6.0155            \\
Back Full-Triple Full-Dou. Full    & bFtFdF             & 5.775            & 6.1215            \\
\bottomrule
\end{tabular}
}
\end{table}

\section{Detailed Scoring Criteria}

\subsection{Scoring Criteria for the Air Stage}
According to the FIS Judges' handbook~\cite{refbook}, the Air stage is evaluated based on the take-off, height, and distance of the competitor's jump. The take-off assessment focuses on how the jump is initiated, while the height and distance are determined by the combination of the athlete's speed and take-off force. Additionally, take-off, height, and distance are evaluated, as well as the length and steepness of the landing hill. The competitor must not land too short (on the knoll) or too long (beyond the transition area of the landing hill). Notably, the height and distance scores are directly derived from competition instruments. The Air stage accounts for 20\% of the overall score (with a maximum of 2 points per judge) and is divided into two components:
10\% Technical Take-Off and 10\% Height and Distance.

Technical take-off refers to how the competitor initiates the jump by extending their body at the precise moment while leaving the kicker. The take-off is judged from when the competitor enters the transition zone until their feet leave the kicker.

\begin{figure*}[!htbp]
\begin{center}
\includegraphics[width=0.75\linewidth]{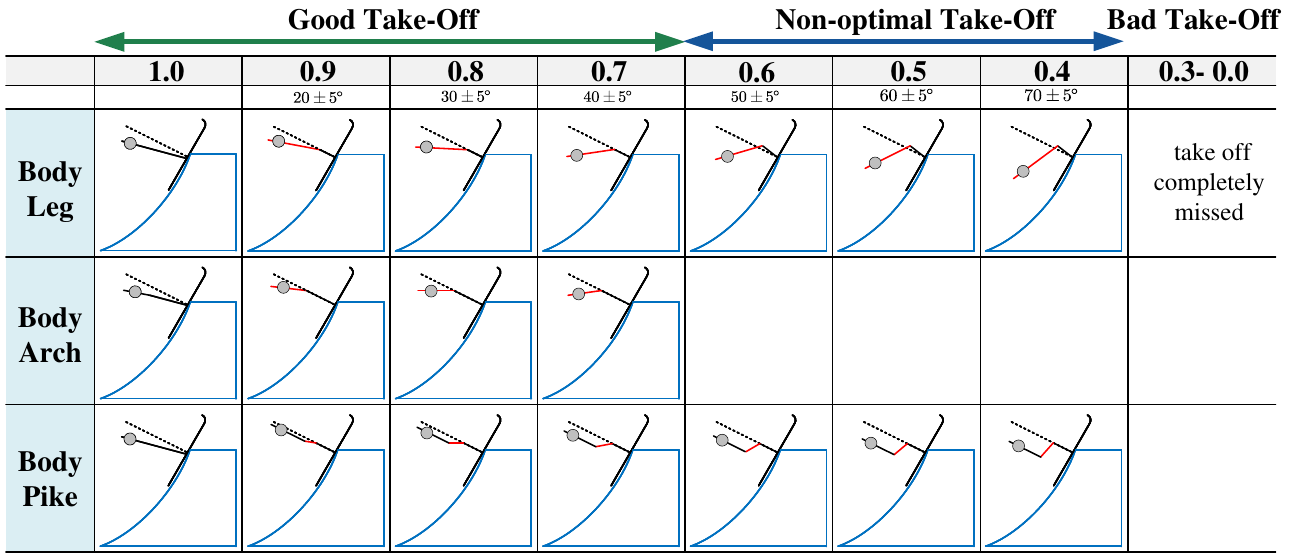}
\end{center}
\caption{\textbf{Detail Deduction Scale (DDS) for Take-Off.} The chart categorizes take-off quality into three main levels: Good Take-Off, Non-optimal Take-Off, and Bad Take-Off, with scores ranging from 1.0 to 0.0. Each row represents a specific body posture during the take-off phase, including Body Leg, Body Arch, and Body Pike.}
\label{fig:8}
\end{figure*}

Figure~\ref{fig:8} provides a detailed visualization of the scoring criteria for the air stage, categorizing take-off quality into three main levels: Good Take-Off, Non-optimal Take-Off, and Bad Take-Off, with scores ranging from 1.0 to 0.0. Each row in the chart corresponds to a specific body posture during the take-off stage, including Body Leg, Body Arch, and Body Pike. Each cell in the figure contains diagrams illustrating the athlete’s posture and corresponding scoring criteria at various take-off angles.

\textbf{Good Take-Off (1.0-0.7):} In this category, the athlete demonstrates optimal posture with minimal angular deviation from the desired trajectory. For example, in the Body Leg row, scores of 0.9 and 0.8 are awarded when the leg alignment and take-off angle fall within acceptable ranges, such as $20^{\circ}\pm 5^{\circ}$ or $30^{\circ}\pm5^{\circ}$.

\textbf{Non-optimal Take-Off (0.6-0.4):} This level indicates significant deviations from ideal angles and noticeable flaws in posture. For instance, in the Body Leg row, take-off angles of $50^{\circ}\pm5^{\circ}$ or $60^{\circ}\pm5^{\circ}$ result in deductions due to visible arching in the athlete's body, which reduces take-off efficiency.

\textbf{Bad Take-Off (0.3-0.0):} This category represents severe take-off failures. The athlete's posture and angle deviate drastically from the norm (\eg, angles exceeding $70^{\circ}\pm5^{\circ}$), or the take-off is entirely missed, resulting in the lowest scores.

\begin{figure*}[!htbp]
\begin{center}
\includegraphics[width=1.0\linewidth]{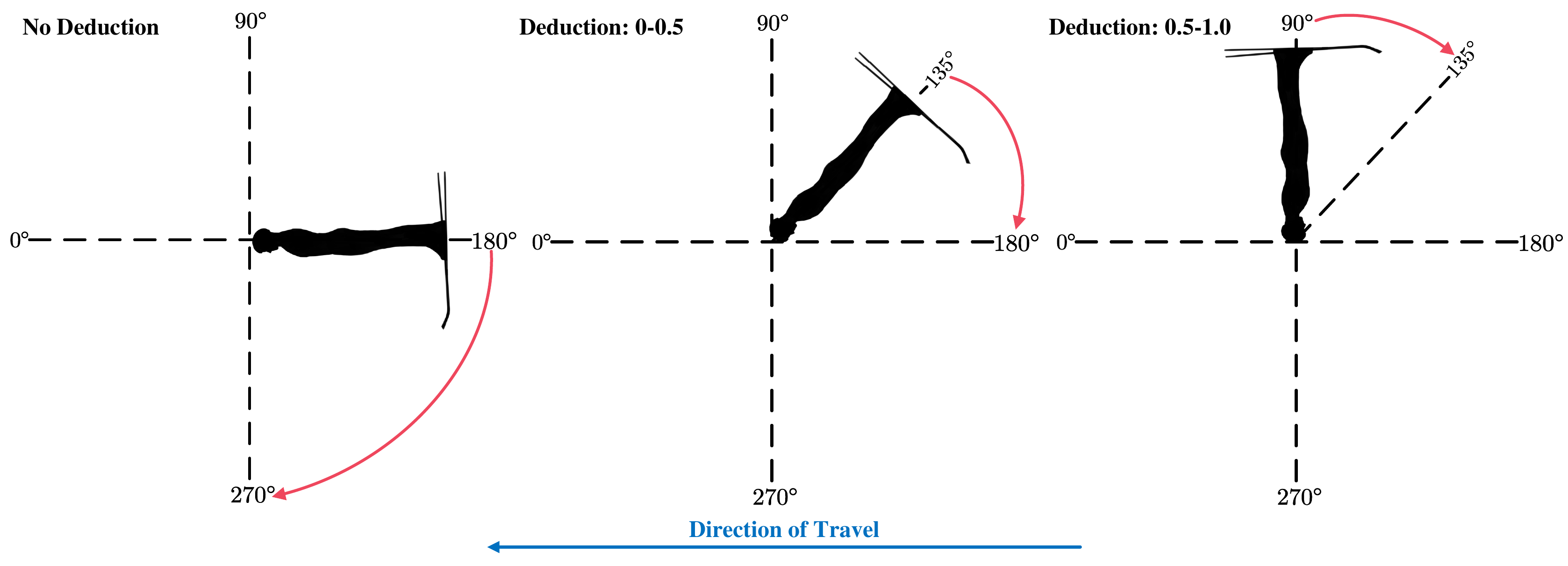}
\end{center}
\caption{\textbf{Deduction rule of early Twist/Tuck/Pike start.} The evaluation is based on when the athlete initiates the movement relative to their aerial position. No deduction will be applied if the twist begins at or after the head and body reach the 9:00 position.}
\label{fig:9}
\end{figure*}

\begin{figure*}[!htbp]
\begin{center}
\includegraphics[width=1.0\linewidth]{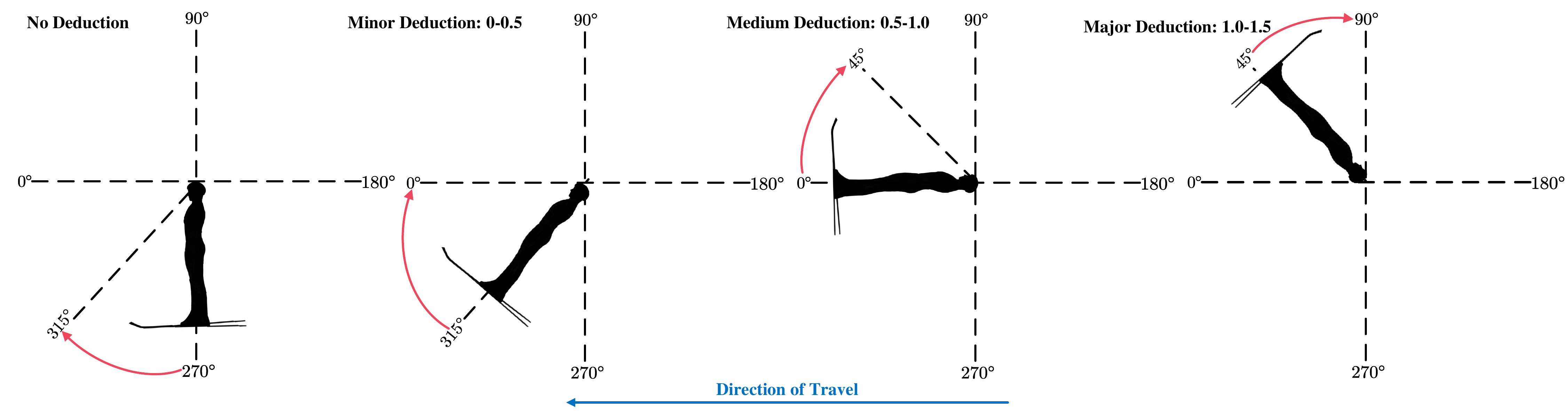}
\end{center}
\caption{\textbf{Deduction Rule of Late Conclusion of Twists in Double Somersaults.} No deduction will be applied if the twist concludes before or exactly at the 9:00 position (90° relative to the athlete’s rotation).}
\label{fig:10}
\end{figure*}

\begin{figure*}[!htbp]
\begin{center}
\includegraphics[width=1.0\linewidth]{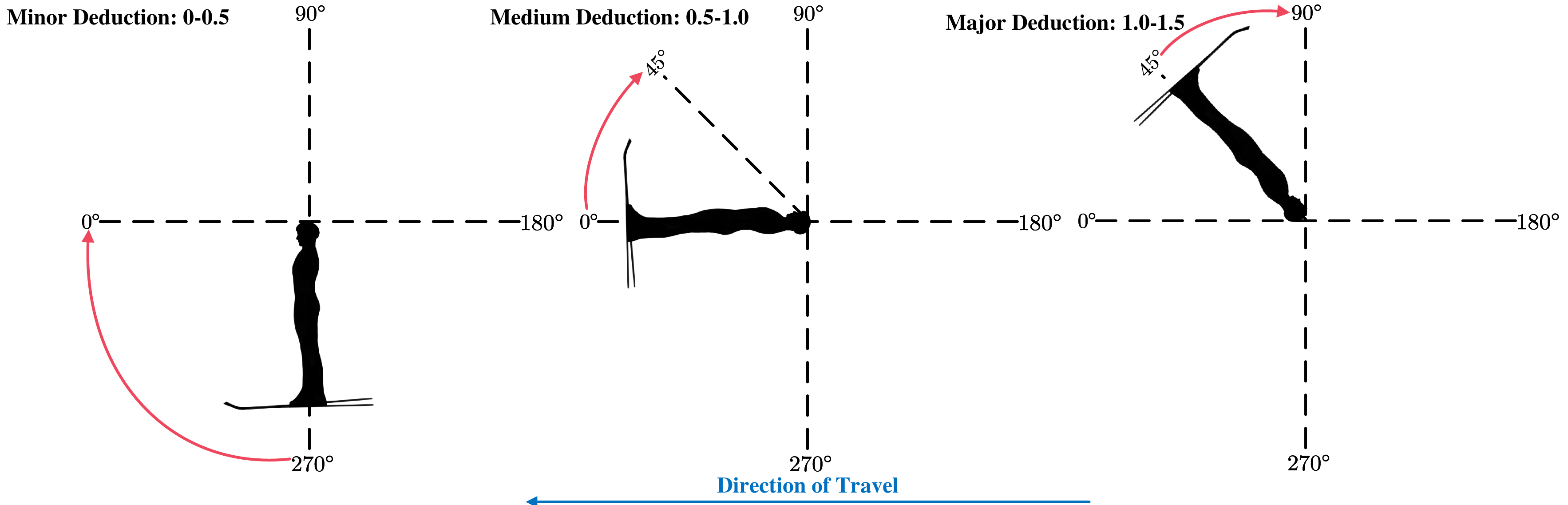}
\end{center}
\caption{Deduction rule of late conclusion of twists in triple somersaults (Twist finish). }
\label{fig:11}
\end{figure*}

\subsection{Scoring Criteria for the Form Stage}
According to the FIS Judges’ handbook~\cite{refbook}, the form stage accounts for 50\% of the overall score (with a maximum of 5 points per judge). In the form stage, we evaluate the athlete’s posture and positioning of the body, skis, arms, hands, and ski poles while in the air. This stage reflects how the athlete executes each maneuver. The evaluation considers several aspects, including precision (e.g., body tightness, economy of movement, balance, and mechanics), aerial stability (or control), separation, and the timing of movements about the take-off. The assessment begins the moment the tips of the skis leave the kicker and continues until the athlete makes contact with the snow.

\subsubsection{Timing}

\textbf{Timing} is a crucial factor in evaluating skiing performance. It refers to an athlete's ability to control the timing of their movements and ensure smooth transitions between different stages of the action. Timing not only influences the fluidity of motion but is also directly tied to the technical quality of execution and the judges’ scoring criteria.

During aerial maneuvers such as flips or twists, Timing determines the flow and coherence of the action. For instance, in a flip, the athlete must precisely manage the intervals between rotations to complete the maneuver within the air stage and adequately prepare for the landing stage. According to FIS scoring standards~\cite{refbook}, Timing plays a role throughout every stage of the performance, ensuring seamless transitions and high-quality execution. Judges evaluate an athlete's technical proficiency by assessing their Timing control.


Figure~\ref{fig:9} illustrates the deduction criteria for initiating a twist, tuck, or pike motion too early during the jump. The evaluation is based on when the athlete begins the movement relative to their aerial position. If the twist begins at or after the athlete's head and body reach the $180^{\circ}$ (relative to the rotation) position, no deduction is applied. This is considered an optimal initiation point that ensures proper execution. If the athlete initiates the twist slightly earlier, such as between the $135^{\circ}$ and $180^{\circ}$ positions, a minor deduction of 0 to 0.5 points is applied. This reflects a premature initiation but with minimal impact on execution quality. If the twist begins much earlier than the $135^{\circ}$ position, the deduction increases to 0.5-1.0 points. This indicates poor timing, which can disrupt the jump’s fluidity and technical precision.

Figure~\ref{fig:10} demonstrates the deduction criteria for finishing a twist too late during double somersaults, focusing on when the twist concludes relative to the aerial rotation. If the twist concludes before or exactly at the $315^{\circ}$ position, no deduction is applied. This indicates a properly timed finish. If the athlete completes the twist slightly late, such as between the $315^{\circ}$ and $0^{\circ}$ positions, a deduction of 0-0.5 points is applied. This reflects minor delays but does not significantly disrupt the overall performance. If the twist concludes even later, such as between the $0^{\circ}$ and $45^{\circ}$ positions, a medium deduction of 0.5-1.0 points is applied, indicating moderate disruption in timing. If the twist concludes much later, such as past the $45^{\circ}$ position, a major deduction of 1.0-1.5 points is applied. This indicates poor control and severely impacts the overall execution quality.

Figure~\ref{fig:11} extends the deduction rules to triple somersaults, focusing on late twist conclusions during these more complex maneuvers. A minor deduction is applied if the twist concludes slightly late, such as after the $270^{\circ}$ position but before the $0^{\circ}$ position. This reflects a small deviation from the ideal timing. A medium deduction is applied if the twist finishes significantly late, such as between the $0^{\circ}$ and $45^{\circ}$ positions. This shows a noticeable delay, impacting the fluidity of the triple somersault. A major deduction is applied if the twist concludes very late, such as past the $45^{\circ}$ position, indicating poor control and execution quality.

\subsubsection{Form Breaks}

The following guidelines are applied when we assess form breaks during a jump:
\begin{itemize}
    \item Minor form breaks deduct up to 25\% of possible form points.
    \item Medium form breaks deduct up to 50\% of possible form points.
    \item Major form breaks deduct up to 100\% of possible form points.
\end{itemize}
For proper execution, the body should remain extended not only during take-off but also prior to landing. The severity of the deduction depends on whether the form could be consistently better throughout the entire jump or only in specific parts of the maneuver. For example, a jump performed with a minor form break during one somersault would result in a small deduction compared to a major form break affecting the entire sequence.


Form in skiing is defined based on specific body positions during the maneuver, which must adhere to the layout position (a straight body with no bending, $0^{\circ}$) or the tuck or pike position (a $90^{\circ}$ bend at the knees and hips). The only exception is the puck position, permitted solely for half-in and twisting front flips. Deductions are assessed based on deviations from the required body position: variations of less than $45^{\circ}$ are considered minor, approximately $45^{\circ}$ are considered medium, and more than $45^{\circ}$ are considered major. These guidelines ensure consistent evaluation of the technical execution of an athlete's form.

\subsubsection{Separation}

Separation refers to the competitor’s ability to demonstrate the beginning and end of each maneuver, such as the declared number of twists within each flip. When a jump includes a different number of twists in each flip, the competitor should distinctly show a change in twisting speed between flips. While the hands can aid in identifying when a twisting maneuver is completed, they are not required to demonstrate separation.
It is important to note that the presence or absence of separation should not significantly impact timing evaluation. A jump may exhibit a clear separation of maneuvers but needs proper timing, and conversely, timing criteria can be satisfied without clear separation. This distinction ensures that separation and timing are assessed as independent criteria.

\subsubsection{Control in Air}

In the form stage, control in the air is crucial, as excessive motion used to adjust speed or balance can lead to form breaks and deductions. Excessive motion may involve pulling or stretching, manifesting in over-rotation or under-rotation during flips. For upright jumps, athletes often rely on arm movements to maintain balance, regulate rotational speed, or adjust their landing trajectory to avoid landing too far forward, backward, or misaligned with the fall line. In the Layout position, flipping speed is adjusted by either pulling the body inward—bending at the knees, waist, or neck—or by stretching the body and extending the arms parallel to the head. Both actions can result in form breaks.

\textbf{NOTE:} During the final flip, once the torso reaches 45 degrees to the horizontal plane, a minor pike (a slight bend at the waist) is acceptable and will not incur a deduction. During this "preparation for landing," athletes may slightly bend at the waist (but not the knees) and spread their legs to shoulder width without being penalized for a form break. However, if the bend at the waist exceeds a minor level (i.e., more than 45 degrees), athletes will be instructed to deduct 0.2 points. All other detail deduction scale values remain applicable throughout the flip until the moment of touchdown.

\begin{figure*}[!htbp]
\begin{center}
\includegraphics[width=1.0\linewidth]{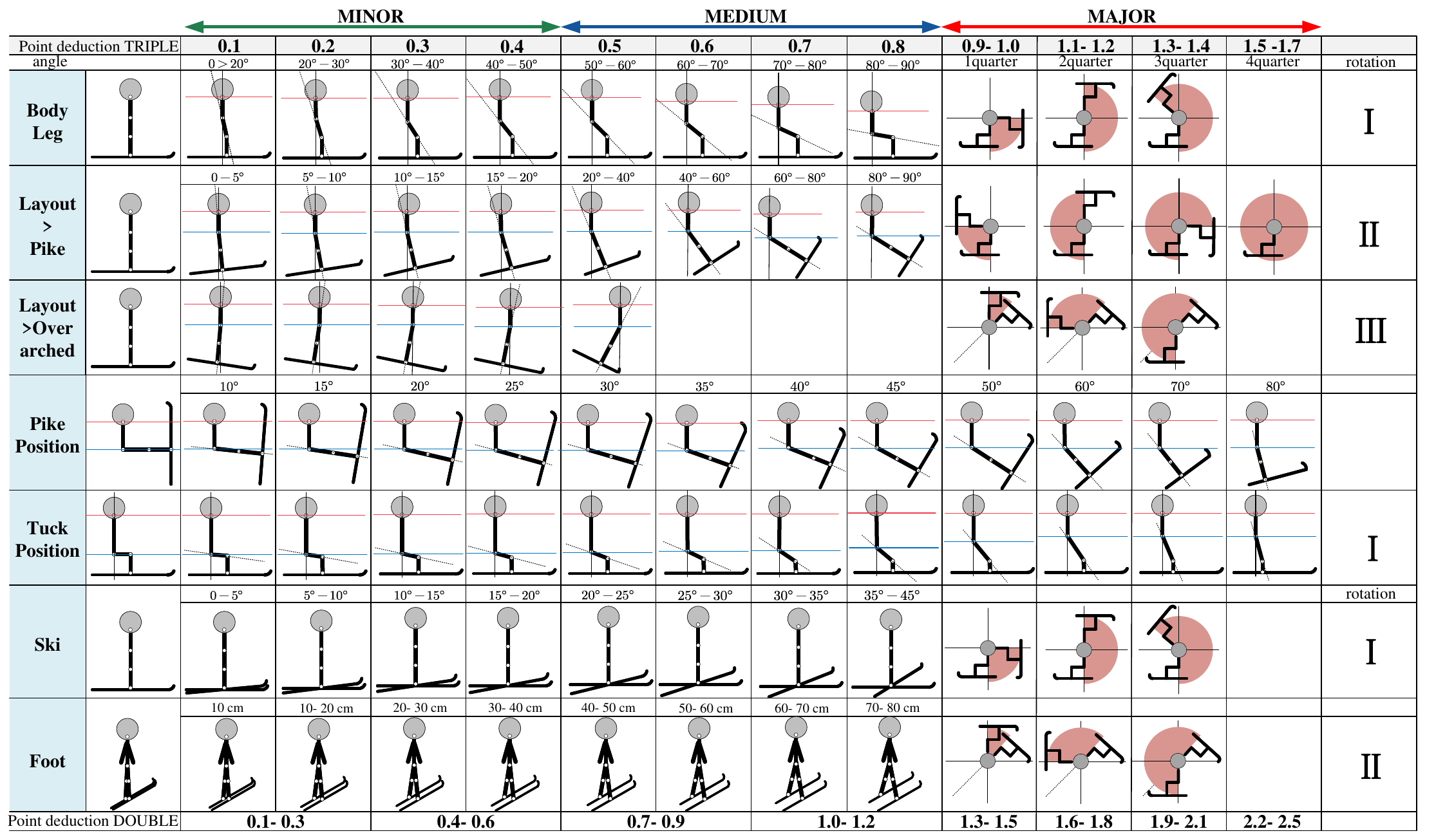}
\end{center}
\caption{\textbf{Detail deduction scale (DDS) for form.} The chart categorizes deductions for form into three main severity levels: Minor, Medium, and Major, with point deductions increasing proportionally. Each row evaluates a specific body position during the form stage, including Body Leg, Layout $>$ Pike, Layout $>$ Overarched, Pike Position, Tuck Position, Ski Alignment, and Foot Placement. Rotational errors and severe form breaks are highlighted in red-marked areas, emphasizing critical mistakes. Scores are adjusted based on the degree of deviation, ensuring precise and consistent scoring.}
\label{fig:12}
\end{figure*}

\subsection{Details}
Figure~\ref{fig:12} provides an in-depth visualization of the scoring criteria for the Form stage in skiing, which evaluates athletes' body alignment, positioning, and specific movement execution during complex aerial maneuvers. The figure systematically categorizes penalties into Minor, Medium, and Major deductions, with associated ranges of point deductions depending on the degree of deviation or error. The critical elements assessed are divided into six rows: Body Leg, Layout $>$ Pike, Layout $>$ Overarched, Pike Position, Tuck Position, Ski, and Foot. Here is a detailed breakdown of each component:


\textbf{Body Leg.} This row focuses on the alignment of the athlete’s body and leg during aerial maneuvers.
\begin{itemize}
    \item Minor Deductions (0.1-0.4 points):
Minor angular deviations between the body and leg, measured in degrees (e.g., $0^{\circ}$-$50^{\circ}$). As the angles increase from $20^{\circ}$ to $50^{\circ}$, deductions also increase, reflecting less precise alignment.
    \item Medium Deductions (0.5-0.8 points):
Moderate misalignment where the angle ranges from $50^{\circ}$ to $90^{\circ}$, indicating noticeable loss of control or lack of tightness in the leg-body form.
    \item Major Deductions (0.9-1.7 points):
Significant form breakdowns, with angles exceeding $90^{\circ}$, indicate severe performance issues. Large rotational deviations, visualized in red, indicate critical errors.
\end{itemize}

\textbf{Layout $>$ Pike.} This row evaluates the transition or improper substitution of a layout form into a pike position:
\begin{itemize}
    \item Minor Deductions (0.1-0.4 points):
Gradual shifts from a clean layout to a slight pike form are penalized based on the angle's degree ($0^{\circ}$-$20^{\circ}$).
    \item Medium Deductions (0.5-0.8 points):
Larger deviations with pike angles reaching $20^{\circ}$-$90^{\circ}$ indicate an apparent loss of form consistency.
    \item Major Deductions (0.9-1.7 points):
Extreme over-piking or form collapse, with angles exceeding $90^{\circ}$, leads to significant deductions. Red-marked areas highlight substantial form breakdowns.
\end{itemize}

\textbf{Layout $>$ Overarched.} This row assesses overarched positions during layout execution:
\begin{itemize}
    \item Minor Deductions (0.1-0.4 points):
Minor over-arch errors (\eg, $0^{\circ}$-$20^{\circ}$) are considered minor but still penalized.
    \item Medium Deductions (0.5-0.8 points):
Moderate overarch ($20^{\circ}$-$40^{\circ}$) with visible loss of clean alignment.
    \item Major Deductions (0.9-1.7 points):
Significant over-arching results in major penalties. These deviations disrupt overall visual appeal and execution quality, marked in red for critical errors.
\end{itemize}

\textbf{Pike Position.} This row focuses on the accuracy of the pike position during execution:
\begin{itemize}
    \item Minor Deductions (0.1-0.4 points):
Slight deviations in pike form, such as angles between $10^{\circ}$ and $25^{\circ}$.
    \item Medium Deductions (0.5-0.8 points):
Noticeable errors with pike angles ranging from $30^{\circ}$ to $45^{\circ}$, indicating poor control.
    \item Major Deductions (0.9-1.7 points):
Severe pike form collapse (exceeding $45^{\circ}$).
\end{itemize}

\textbf{Tuck Position.} This row evaluates the tuck position, particularly its compactness and accuracy:
\begin{itemize}
    \item Minor Deductions (0.1-0.4 points):
Slight deviations in tuck angles ($10^{\circ}$-$25^{\circ}$) are penalized.
    \item Medium Deductions (0.5-0.8 points):
Moderate tuck position issues, with angles increasing to $30^{\circ}$-$45^{\circ}$, indicating loose or poorly controlled tucks.
    \item Major Deductions (0.9-1.7 points):
Significant tuck errors, with angles exceeding $45^{\circ}$, leading to severe deductions.
\end{itemize}

\textbf{Ski.} This row focuses on the alignment and positioning of the athlete’s skis during the maneuver:
\begin{itemize}
    \item Minor Deductions (0.1-0.3 points):
Minor spacing errors between skis, such as $0^{\circ}$-$20^{\circ}$ of separation.
    \item Medium Medium Deductions (0.4-0.6 points):
Moderate spacing issues, with ski separation ranging from $20^{\circ}$-$45^{\circ}$, indicating alignment issues.
    \item Major Deductions (0.7-0.9 points):
Severe ski misalignment, with separation exceeding $45^{\circ}$, leading to significant deductions.
\end{itemize}

\textbf{Foot.} This row evaluates the positioning of the athlete’s feet during the maneuver:
\begin{itemize}
    \item Minor Deductions (0.1-0.3 points):
Slight misplacement, such as minor deviations in the foot position.
    \item Medium Medium Deductions (0.4-0.6 points):
Clear positioning issues with noticeable foot misalignment.
    \item Major Deductions (0.7-1.2 points):
Substantial errors, including excessive foot misplacement, disrupt the form.
\end{itemize}

In summary, Fig.~\ref{fig:12} illustrates the three scoring dimensions for the form stage: Penalty Severity, Rotational Penalties, and Detailed Criteria. Penalty Severity categorizes deviations into three levels, \ie, minor, medium, and major, with increasingly higher deductions for more severe errors. Rotational Penalties focus on mistakes involving rotation, with red-marked areas indicating critical form breakdowns. Detailed Criteria provide a granular assessment of specific body positions in each row, ensuring accurate and consistent scoring across different performances.

\begin{figure}[!htbp]
\begin{center}
\includegraphics[width=1.0\linewidth]{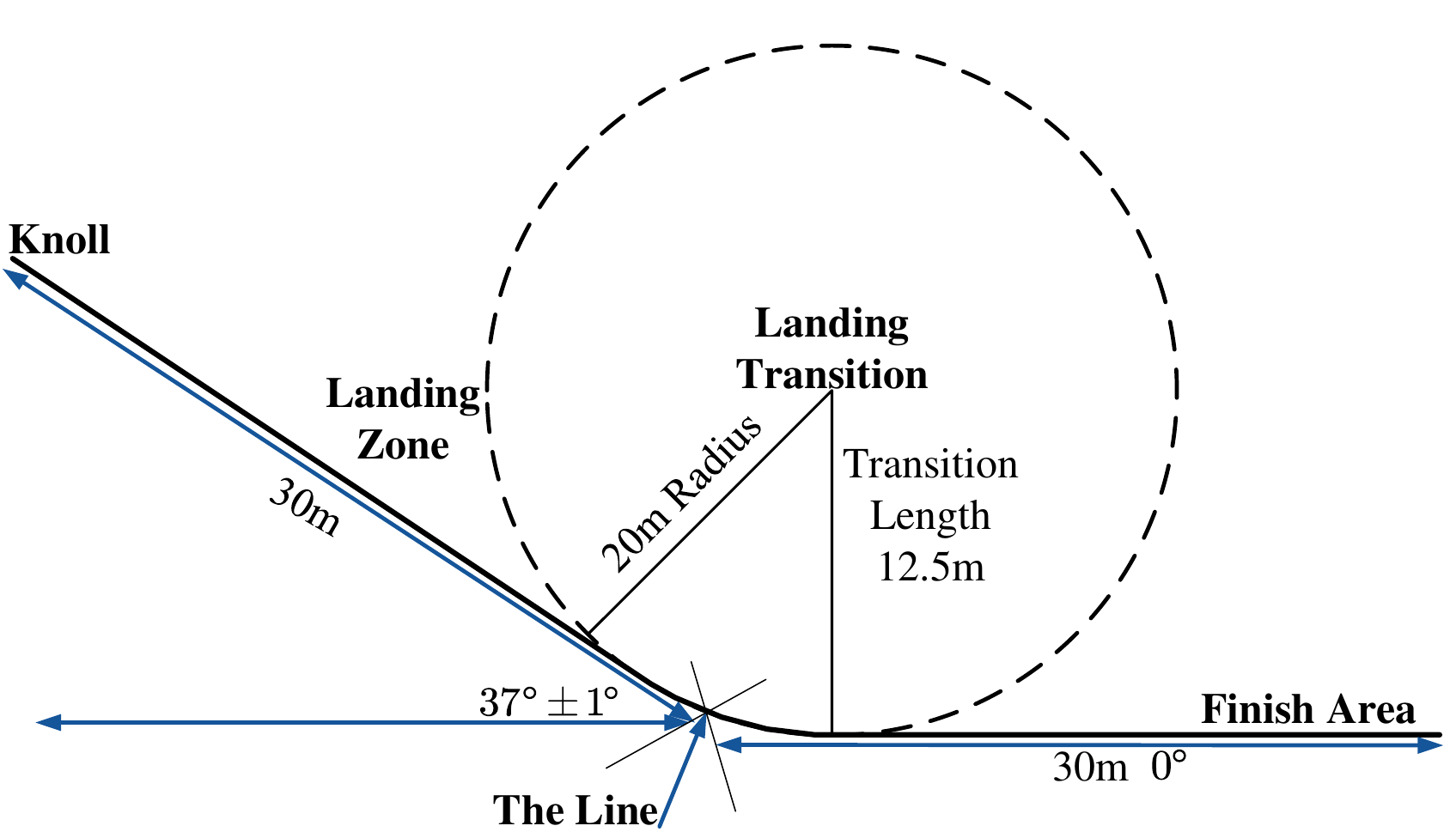}
\end{center}
\caption{\textbf{The key areas of the landing phase in skiing}. Including the complete layout from the Landing Zone to the Transition Zone and the Finish Area. }
\label{fig:13}
\end{figure}

\subsection{Scoring criteria for the Landing Stage}
According to the FIS Judges' handbook~\cite{refbook}, a proper landing requires a balanced, stable, and controlled body position throughout the entire stage. The competitor should exhibit precision and grace, ensuring minimal disruption upon contact with the landing surface. Absorption of impact should primarily involve the knees and lower body, with only a slight bend at the waist. The evaluation of the landing begins when the competitor makes contact with the snow and continues until they demonstrate sufficient skiing control as they transition from the landing hill to the finish area. The landing stage accounts for 30\% of the total score, with a maximum of 3.0 points per judge.

Figure~\ref{fig:13} illustrates the critical areas of the landing stage in skiing, spanning the Landing Zone, Transition Zone, and Finish Area. The landing stage is critical following the execution of aerial maneuvers, where the athlete transitions from the Knoll region to the landing track. Typically characterized by a steep slope, this region is designed to ensure sufficient height and speed before landing.

The Landing Zone is the primary area for landing, measuring 30 meters in length with a slope angle of $37^{\circ}\pm 1^{\circ}$. The athlete’s objective is to land stably within this designated range. This zone is crucial for scoring, as judges evaluate whether the athlete makes safe and steady contact with the ground. Any loss of balance or deviations from expected movements during landing may result in penalties. Furthermore, failure to transition smoothly in the Landing Transition Zone, \eg, falling or sliding unstably, will also negatively affect the score. Judging criteria for the landing stage include specific deductions based on contact with the snow:
\begin{itemize}
    \item \textbf{Hand Contact.} A maximum score of 2.0 can be awarded.
    \item \textbf{Body Contact.} A maximum score of 1.5 can be awarded.
    \item \textbf{No Touch.} Even without hand or body contact, deductions may occur for severe imbalances, skiing sideways, circling, or skiing backward.
\end{itemize}

The landing is evaluated from when the athlete makes contact with the snow until they transition from the landing hill to the finish area. A poor landing does not necessarily result in deductions from the form score, as form deductions are based solely on specific errors during the aerial maneuver. Mistakes during the landing will only affect the landing score. A well-executed jump will still receive an appropriate form score, with landing deductions reflecting errors explicitly made during the landing stage.

\backmatter

\bibliography{ref}

\end{document}